\documentclass[review]{elsarticle}
\usepackage{amsmath}   
\usepackage{lineno} 
\usepackage{hyperref}
\usepackage{algorithm}
\usepackage{algpseudocode}
\usepackage{bbding}
\usepackage{pifont}
\usepackage{wasysym}
\usepackage{amssymb}

\usepackage{amssymb}   
\usepackage{booktabs}   
\usepackage{multicol}  
\usepackage{multirow}  
\usepackage{diagbox}    
\usepackage{multirow}   
\usepackage{bm}  
\journal{Journal of \LaTeX\ Templates}   

\usepackage{graphicx} 
\usepackage{caption} 
\usepackage{subfigure} 
\usepackage{float}  
\usepackage[subfigure]{graphfig} 

\begin{document}

\begin{frontmatter}

\title{Graph-based Multi-view Binary Learning for Image Clustering}

\cortext[cor1]{Corresponding author: Xianping Fu}
\author[1]{Guangqi Jiang}
\author[1]{Huibing Wang}
\author[1]{Jinjia Peng}
\author[1]{Dongyan Chen}
\author[1,2]{Xianping Fu\corref{cor1}}

\address[1]{College of Information and Science Technology, Dalian Maritime University, Danlian, Liaoning, 116021, China}
\address[2]{Pengcheng Laboratory, Shenzhen, Guangdong, 518055, China}

\begin{abstract}    

Hashing techniques, also known as binary code learning, have recently gained increasing attention in large-scale data analysis and storage. Generally, most existing hash clustering methods are single-view ones, which lack complete structure or complementary information from multiple views. For cluster tasks, abundant prior researches mainly focus on learning discrete hash code while few works take original data structure into consideration. To address these problems, we propose a novel binary code algorithm for clustering, which adopts graph embedding to preserve the original data structure, called (Graph-based Multi-view Binary Learning) GMBL in this paper.  GMBL mainly focuses on encoding the information of multiple views into a compact binary code, which explores complementary information from multiple views. In particular, in order to maintain the graph-based structure of the original data, we adopt a Laplacian matrix to preserve the local linear relationship of the data and map it to the Hamming space. Considering different views have distinctive contributions to the final clustering results, GMBL adopts a  strategy of automatically assign weights for each view to better guide the clustering. Finally, An alternating iterative optimization method is adopted to optimize discrete binary codes directly instead of relaxing the binary constraint in two steps. Experiments on five public datasets demonstrate the superiority of our proposed method compared with previous approaches in terms of clustering performance.
\end{abstract}

\begin{keyword}    
\texttt  Multiview  \sep Binary code \sep Clustering \sep Graph-based
\end{keyword}

\end{frontmatter}



\section{Introduction}   

With the development of computer vision applications, we have witnessed that hash technology has become an indispensable step in the processing of large data \cite{wang2015learning} \cite{bernabe2019efficient}. In dealing with data analysis, organization, and storage, etc., there is an imminent need to use the effective hash code to process data clustering from big databases. Besides, most existed digital devices mainly based on binary code, which can effectively save computing time and storage space. In general, the similarity between the original data can be effectively preserved by encoding the original high-dimensional data using a set of compact binary codes \cite{dean2013fast}, \cite{fuentes2019topic}. These advantages make it obtained widely applied in the computer vision task, such as image clustering \cite{sudharshan2019multiple}, image retrieval \cite{ahmed2019hash} and multi-view learning \cite{yang2017discrete} etc.

Nowadays, binary coding methods have been well investigated in many fields. Locality Sensitive Hashing (LSH) \cite{datar2004locality} pioneered hash research by indexing similar data with hash codes and achieved large-scale search in constant time. Commonly, the hashing method can be roughly divided into two major categories: supervised model and unsupervised model.  The supervised hash code generates a discrete, efficient and compact hash code by using the label information of the data. For instance, Minimal Loss Hashing (MLH) \cite{norouzi2011minimal}, Supervised Discrete Hashing (SDH) \cite{shen2015supervised}, Supervised Discrete Hashing With Relaxation (SDHR) \cite{gui2016supervised} and Fast Supervised Discrete Hashing (FSDH) \cite{gui2017fast}. However, the problem of manually labeling large-scale data are very expensive has not been considered. Thus, the unsupervised hash method is proposed to address this problem, which also obtained good performance in binary code learning. Unsupervised hash models include, but not limited to, Spectral Hashing (SH) \cite{weiss2009spectral}, Iterative Quantization (ITQ) \cite{gong2012iterative}, Discrete Graph Hashing (DGH) \cite{liu2014discrete}, inductive Hashing on Manifolds \cite{shen2015hashing} etc.. Because discrete hash codes reduce the quantization error, Discrete Hash (DGH) \cite{liu2014discrete} and Supervised Discrete Hash (SDH) \cite{shen2015supervised} have significant improvements in hash coding performance. 

Up to now, most methods usually use a single view to learn binary code representation, which fails to explain the observed fact that the complementary features and diversity of multiple views. In many visual applications\cite{wang2017effective}, \cite{wu2018deep} \cite{wang2015unsupervised} \cite{wu2016exploiting} , data is usually collected from datasets in various fields or from different feature extractors\cite{wu2018deep1,wang2016iterative}, such as Histogram \cite{dalal2005histograms}, Local Binary Patters (LBP) \cite{ojala2002multiresolution} and Scale Invariant Feature Transform (SIFT) \cite{rublee2011orb} etc.. Compared with single-view information, multi-view data maybe include more potential comprehensive information. Therefore, multi-view learning obtained more and more attention in many applications. Xia et al. \cite{xia2010multiview} introduced a spectral-embedding algorithm to explore the complementary information of different views, which have proved effective for image clustering and retrieval. Zhang et al. \cite{zhang2016flexible} explicitly produced low-dimensional projections for different views, which could be applied to other data in the out-of-sample. Wang et al. \cite{wang2015robust} effectively maintain well-encapsulated individual views while study subspace clustering for multiple views. Therefore, gathering information from multiple views and exploring the underlying structure of data is a key issue in data analysis. In addition, since the hash method could efficiently encode high-dimensional data, a promising research field by adopting multi-view binary code to improve clustering performance.

Recently, some efforts have been done to learn effective hash code from multi-view data \cite{wu2018cycle}. There are two types of research areas: cross-view hashing and multi-view hashing. Song et al. proposed a novel Imedia Hashing Method (IMH) method, which can explore the relevance among different media types from various data sets to achieve large-scale retrieval inter-media. Besides, Zhu et al. \cite{zhu2013linear} proposed Linear Cross-modal Hashing (LCMH) has obtained good performance in cross-view retrieval tasks. Ding et al. \cite{ding2014collective} by using the latent factor models from different modalities collective matrix decomposition. Composited Hashing with Multiple Information Sources (CHMIS) \cite{zhang2011composite} is the first work in the multi-view hash field. More recently, Multiview Alignment Hashing (MAH) \cite{liu2015multiview} based on nonnegative matrix factorization can respect the distribution of data and reveal the hidden semantics representation. Then many multi-view hash methods are proposed, such as Discrete Multi-view Hashing (DMVH) \cite{yang2017discrete} and Multi-view Discrete Hashing (MvDH) \cite{shen2018multiview}. Most of these related works of hash focus on mutual retrieval tasks between different views, which ignored the potential cluster structure and distribution of information in multi-view data. Therefore, hash technology is of vital significance for multi-view clustering and arouses attention from researchers in the computer vision region. Table \ref{table 1} summarizes the current multi-view hash methods from model learning paradigms, hash optimization strategies, and categories.


\begin{table}[]
	\scriptsize
	\centering
	\caption{Comparison of several multi-view hash algorithms}   	
	\label{table 1} 
	\begin{tabular}{ccccccccccc}
		\hline
		Methods            & IMH & LCMH & CMFH	& CHMIS  & MAH & DMVH & MvDH & Ours\\ \hline
		Multi-view cluster & \ding{56}  &\ding{56} & \ding{56} & \ding{56}  & \ding{56} & \ding{56} & \ding{56} &  \Checkmark \\
		Discrete           & \ding{56} & \ding{56}  &\ding{56} &\ding{56}  &\ding{56} & \Checkmark & \Checkmark & \Checkmark \\
		Unsupervised       & \ding{56}  & \Checkmark  & \Checkmark & \Checkmark  & \Checkmark &\ding{56} & \Checkmark & \Checkmark \\    \hline   
	\end{tabular}
\end{table}

\begin{figure}[htb]
	\centering	
	\includegraphics[width=13.5cm]{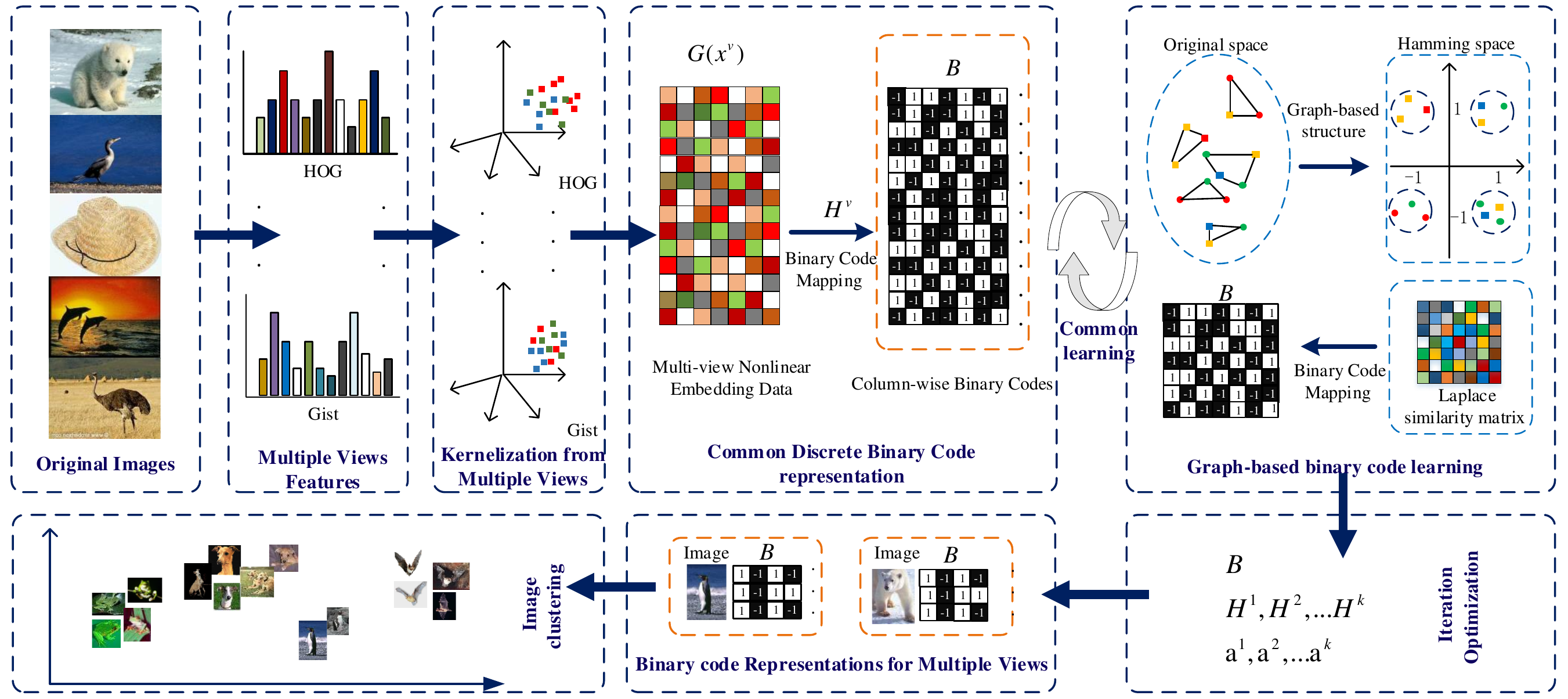}
    \caption{The Construction Process of GMBL}	
    \label{Fig.1} 
\end{figure}

In this paper, we introduce a novel frame for graph-based multi-view binary code clustering. In order to learn an efficient binary code, our method attempts to efficiently learn discrete binary code and maintain manifold structure in Hamming space for multi-view clustering tasks. To learn discriminated binary codes, the key design is to generate similar binary codes for similar data without destroying the inherent attributes of the original space, which can share information between multiple views as much as possible. By learning the cooperative work between hash codes and graph, clustering tasks and coding efficiency is significantly improved. Since direct optimization of binary codes is a difficult problem, an effective alternating iterative optimization strategy is developed to solve the hash coding. The construction process of GMBL has been shown as Fig.\ref{Fig.1}. The main contributions of this paper are illustrated as follows:

\begin{itemize} 

\item We propose an innovative unsupervised hashing method to learn compact binary codes from multi-view data. To preserve the original structure of input data, our proposed method combines hash codes learning and graph clustering through Locally Linear Embedding learning. Joint learning ensures that the generated hash code can improve performance for clustering. 

\item Inspired in graph learning, the local similar structure of the original information is embedded into the Hamming space to learn compact hash codes. The view-specific information is shared from multiple views by projecting different views of original features into the common subspace through local linear embedding. 

\item In order to obtain an accurate clustering effect, we assign different weights to various views according to contribute for clustering. In addition, we introduce the alternating optimization algorithm with strict convergence proof in a new discrete optimization scheme to solve hash coding. 	

\end{itemize}

\section{Related work}    

Most hashing algorithms are based on single-view data to generate binary codes. In this section, we first introduce theories and notations of multi-view binary code learning. Then, we review a classical spectral embedding method by graph Laplacian matrix to preserve the data similarity structure. We will present how to learn binary code from multiple views, and then study complementary hash codes with similarity preservation in the next section.
\subsection{Binary Code Learning}  
 Assume we are given a dataset of $N$ examples $\left\{ {\bm{x}_i^v} \right\}_{i = 1}^N$ with $v$th views. The multi-view matrix in $v$th view can be represented as: ${\bm{X}^v} = \left\{ {\bm{x}_1^v,...,\bm{x}_n^v} \right\} \in {\Re^{{d_v} \times N}}$.

Where ${{d_v}}$ is the dimension of $v$th view. Unsupervised hashing to map the high-dimensional data ${\bm{x}_i} \in {\Re^d}$ into the binary codes  $ {\bm{b}_i}\in \left\{ {-1,1} \right\}$. Therefore, binary code generating is to learn a set of hash project functions to produce the corresponding set of $l$-bits binary code. For the $v$th views sample of hash function as: ${{\rm\bm{H}}^v} = \left\{ {\bm{h}_j^v( \cdot )} \right\}_{j = 1}^l$,
Where $ \bm{h}_j^v( \cdot ):{\Re^{{d_v}}} \to \left\{ {-1,1} \right\} $ is a binary mapping. Such functions are usually constructed by combining dimension reduction and binary quantization. Since Hamming distance represents the similarity between binary codes, the hash objective function in $v$th view can be constructed. Then, the binary code of the $v$th view dataset can be written as:
\begin{equation}{\bm{B}^v}=[\bm{h}_1^v({\bm{x}^v}),...,\bm{h}_l^v({\bm{x}^v})] \end{equation}

Where ${\bm{B}^v} \in {\left\{ { - 1,1} \right\}^{{l_v} \times N}}$ is the corresponding hash code of the whole dataset. In the process of binary code mapping, it is necessary to minimize the loss of data and the destruction of the original structure.

\subsection{Single-view Graph Learning} 

The main purpose of similarity preservation is to preserve the geometry structure of manifold data by the local neighborhood of a point, which can be efficiently approximated by the point nearest neighbors. Generally, it has two steps, i.e., discovering similar neighbors and constructing weight matrix.

Let $ {\bm{X}} = \left\{ {\bm{x}_1,...,\bm{x}_N} \right\} $ denote a feature of the samples and  $ {\bm{Y}} = \left\{ {\bm{y}_1,...,\bm{y}_N} \right\} $ denotes $\bm{y}_i$ is a low dimensional vector mapped from $\bm{x}_i$. Firstly, each sample $\bm{x}_i$ is approximated to its $k$-nearest neighbor samples. Then minimizes the reconstruction error in original space are be used as follows:

\begin{equation}
\begin{array}{l}\mathop {\min }\limits_P \sum\limits_{i = 1}^N {\left\| {{x_i} - \sum\limits_{j = 1}^N {{\bm{P}_{ij}}{x_j}} } \right\|} {\rm{ ,  }} \\
s.t.\sum\limits_{i = a}^N {{\bm{P}_{ij}} = 1}\end{array} 
\end{equation}

Where $\bm{P}_{ij}=0$ if $\bm{x}_i$ and $\bm{x}_j$ are not neighbours. Local liner embedding assumes that such linear combinations should remain unchanged even if the manifold structure is mapped to a lower space. Then, used the low-dimensional representation minimizes the reconstruction error as follows:

\begin{equation}
\begin{array}{l}
\mathop {\min }\limits_Y {\sum\limits_i {\left\| {{\bm{y}_i} - \sum\limits_j {{\bm{P}_{ij}}{\bm{y}_j}} } \right\|} ^2} = tr\left( {{\bm{Y}^T}\bm{L}\bm{Y}} \right),\\
s.t.\sum\limits_i {{\bm{y}_i}}  = 0,{\rm{ }}\frac{1}{\bm{N}}\sum\limits_i {{\bm{y}_i}} {\bm{y}_i}^T = {\bm{I}}
\end{array}
\end{equation}
Where $\bm{Y}$ is a low dimensional matrix mapped from $\bm{X}$. Where  $\bm{L} = {(\bm{I} - \bm{P})^T}(\bm{I} - \bm{P})$ is the graph Laplacian matrix and $tr\left(  \cdot  \right)$ is the trace of matrix. 


\section{Graph-based Multi-view Binary Learning}

In this section, we first propose a novel clustering method called Graph-based Multiview Binary Learning(GMBL), which map the data to Hamming space and implement clustering tasks by efficient binary codes. Firstly, the anchor points of data are selected randomly, and the different views are mapped to the same dimension by nonlinear kernel mapping in section 3.1. Then, we propose a method of mapping hash codes, which can learn efficient binary codes with balanced binary code distribution in section 3.2. Furthermore, similarity preservation of different views means that similar data will be mapped to binary code by a short Hamming distance. To do this, our proposed method preserves the local similar structure of data through a similar matrix in section 3.3. Finally, an alternating iterative optimization strategy is applied to search for the optimal solution and the optimization process is illustrated in describe in section 3.4.
		
Suppose our multi-view dataset can be represented ${X^v} = \left\{ {x_1^v,x_1^v...x_N^v} \right\} \in {\Re^{{d_v} \times N}}$, where ${X^v} \in {\Re^{{d_v} \times N}}$ contains all features matrix from the $v$th view, $\bm{d}_v$ is the corresponding feature dimension and $N$ is the total number of samples. The aim of our method is to learn hash code $B \in {\left\{ { - 1,1} \right\}^{r \times N}}$ to represent multi-view data, where $r$ is the binary code length. And some important formula symbols are summarized as follows in table \ref{table 2}.

\begin{table}[htb]
    \scriptsize	
	\centering	
	\caption{The description of important formular symbols}   
	\label{table 2}
	\begin{tabular}{c|l} 
		\hline 
   \textbf{Notation} & \textbf{Description}                                                    \\
   		\hline 
     ${\bm{X}^v}$ & Feature matrix of the $v$th view data                           \\
     $\bm{G}\left( {{\bm{x}^v}} \right)$& Encode each feature vector.of the $v$th view                    \\
    ${\bm{H}^v}$ & Map matrix for features in the $v$th view                       \\
    ${\bm{B}}$   & Collaborative binary code matrix                               \\
     ${\bm{b_i}}$&The hash code representation to the $i$th sample                 \\    
    $\bm{o}_i^v$ & The anchor samples from the $v$th view                          \\
    ${\bm{a}^v}$ & The weighting factor for the $v$th view                         \\
    ${\bm{L}}$   & Set of all features Laplace matrix                             \\
    $\bm{s}_i^v$ & The spares relationships for the i-th feature in the $v$th view \\
	$\bm{d}_v$   & The dimension of features in $v$th view          \\
	${\bm{u}_g^v}$ & $g$th-nearest points in $\bm{G}\left( {{\bm{x}^v}} \right)$ with the $v$th view \\
	     \hline			
		
	\end{tabular}
\end{table}

\subsection{Kernelization from Multiple Views}
We normalize the data from each view to maintain the balance of the data. Since the dimensions of different views may be various, we demand to find an effective method to embed multi-view data into a low-dimensional representation space.

In order to obtain low-dimensional representation, GMBL adopts nonlinear kernel mapping for each view. Inspired by \cite{zhang2018binary} the simple nonlinear RBF kernel mapping method was used to encode each feature vector. GMBL adopt the above technique to explore various information for each view as follows: 
\begin{equation}
                \bm{G}(\bm{x}_i^v) = {\left[ {{{\exp ( - {{\left\| {\bm{x}_i^v - \bm{o}_1^v} \right\|}^2}} \mathord{\left/
				{\vphantom {{\exp ( - {{\left\| {\bm{x}_i^v - \bm{o}_1^v} \right\|}^2}} \xi }} \right.
				\kern-\nulldelimiterspace} \xi }),...,{{\exp ( - {{\left\| {\bm{x}_i^v - \bm{o}_s^v} \right\|}^2}} \mathord{\left/
				{\vphantom {{\exp ( - {{\left\| {\bm{x}_i^v - \bm{o}_s^v} \right\|}^2}} \xi }} \right.
				\kern-\nulldelimiterspace} \xi })} \right]^T}
\end{equation}

Where $\xi$ is the kernel width, and $\left\{ {\bm{o}_j^v} \right\}_{j = 1}^{\rm{s}}$ are the $s$ anchor points are randomly selected from the $v$th view. In the algorithm, we choose the number of anchor points to mapping based on the size of the dataset. Besides, projecting data into the kernel space can avoid the problem of uneven dimensions. $\bm{G}(\bm{x}_i^v) \in {\Re ^s}$ represents the $s$-dimensional nonlinear embedding of data features from the $v$th view $\bm{x}_i^v \in {\Re^{{d_v}}}$. 


\subsection{Common Discrete Binary Code representation}

The features of different views are mapped into hash codes in Hamming space by the projection matrix. 
The representation of the hash code is $(\bm{G}(x_i^v);{\bm{H}^v}) = {\mathop{\rm sgn}} ({\bm{H}^v}\bm{G}(\bm{x}_i^v)) \in {\left\{ { - 1,1} \right\}^{r \times n}}$,where $\bm{b}_i$ is the common binary code representation of the $i$-th sample from different views. ${\mathop{\rm sgn}} \left(  \cdot  \right)$ is a sign operator function. ${\bm{H}^v} \in {\Re ^{r \times s}}$ is the projection matrix of the $v$th view. GMBL combines different views to embed them simultaneously into a common Hamming subspace. The purpose of our method is learning an efficient projection matrix ${\bm{H}^v} $, which to map all samples in the original space into binary code. Therefore, we construct a minimizing loss function as follows:
\begin{equation}
\mathop {\min }\limits_{{\bm{H}^v},{\bm{b}_i}} \sum\limits_{v = 1}^k {\sum\limits_{i = 1}^N {\left\| {{\bm{b}_i} - {\bm{H}^v}\bm{G}(\bm{x}_i^v)} \right\|_F^2} }
\end{equation}

Here ${\bm{b}_i}$ is a binary code for the $i$th samples. By optimizing the above formula, we can get an efficient binary code. It is important to note that learning equilibrium and stable binary codes by using regularized item constraints. In general, using the maximum entropy principle that the equation can be rewritten as:

\begin{equation}
\begin{aligned}
\mathop {\min }\limits_{{\bm{H}^v},{\bm{b}_i},{\bm{a}}} &\sum\limits_{v = 1}^k {{{({a^v})}^c} 	\sum\limits_{i = 1}^N {\left\| {{\bm{b}_i} - {\bm{H}^v}\bm{G}(x_i^v)} \right\|_F^2} }  + \delta \left\| {{\bm{\bm{H}}^v}} \right\|_F^2 \\
- \lambda &\sum\limits_{{\rm{i}} = 1}^N {\left( {{\bm{H}^v}\bm{G}(\bm{x}_i^v){{\left( {{\bm{H}^v}\bm{G}(\bm{x}_i^v)} \right)}^T}} \right)} \\
s.t.{\rm{ }}&\sum\limits_v {{\bm{a}^v}}  = 1,{\bm{a}^v} > 0,{\bm{b}_i}{\left\{ { - 1,1} \right\}^{r \times 1}}
\end{aligned}
\end{equation}

Where ${\bm{a}^v}=[{a}^1,{a}^2,...{a}^k]$ is a nonnegative normalized weighting vector assigned according to the contributes of different views. $c$>1 is the weight parameter, which ensures all views have a special contribution to the final low-dimentional representation. The first item of the equation ensures to learn efficient binary code for multi-view data. The last two terms of the equation are constraints for learning binary code. In this way, adding regularization on $\bm{B}$ can ensure a balanced partition and reduce the redundancy of binary codes. 

\subsection{Graph-based binary code learning}
This section introduces the method of similarity preservation for mapping data to binary codes. Due to the existence of similar underlying structures in different views, the structural features of the original data should also be considered when learning the binary code projection matrix. Keeping the similarity of data is one of the key problems of the hashing algorithms, which means that similar data should be mapped to binary codes with short Hamming distance. Based on this problem, we propose a method to construct a similarity matrix, which can not only preserve the local structure of the data but also preserve the similarity between the data. Then, we introduce the similarity preservation method to map data into binary codes.


In many graph-based hash methods, a key step in similarity preservation is to build neighborhood graphs on the data. For the $v$th view of each data point, we pick up all points set $\left\{ {\bm{u}_g^v} \right\}_{g = 1}^G$ from $\bm{G}(x_i^v)$ to  reconstruct $\bm{G}(x_i^v)$. Where ${\bm{u}_g^v}$ is one of $G$-nearest points. Thus, The optimization equation can be obtained as follows:

\begin{equation}
\begin{array}{l}
\mathop {\min }\limits_{{w^v}} \sum\limits_{i = 1}^N {\left\| {\bm{G}(x_i^v) - \sum\limits_{g = 1}^G {\bm{w}_{ig}^v\bm{u}_g^v} } \right\|} _F^2\\
s.t.{\rm{ }}\sum\limits_{g = 1}^G {\bm{w}_{ig}^v}  = 1 \label{similar}
\end{array} 
\end{equation}
By solving Eq.(\ref{similar}), we get 

\begin{equation}\bm{w}_{ig}^v = \frac{{\sum\nolimits_{t = 1}^G {\left( {\bm{C}_{gt}^{ - 1}} \right)} }}{{\sum\nolimits_{p = 1}^G {\sum\nolimits_{q = 1}^G {\bm{C}_{pq}^{ - 1}} } }}\end{equation}

Where ${\bm{C}_{gt}} = {(\bm{G}(x_i^v) - \bm{u}_g^v)^T}(\bm{G}(x_i^v) - \bm{u}_t^v)$, $\bm{C}$ is a covariance matrix. Where $\bm{w}^v$ are described the relationship between data points, which we can use to define the similar matrix as:

\begin{equation}\bm{S}_{ij}^v = \left\{ \begin{array}{l}
\bm{w}_{ig}^v{\rm{  }} \quad if {\rm{ }} \ x_j^v{\rm{ }}\ is{\rm{ }}\ a{\rm{ }}\ neighbor{\rm{ }}\ of{\rm{ }}\ x_i^v{\rm{ }}\ in{\rm{ }}\ \bm{Q}\left( {x_i^v} \right)\\
0{\rm{     }}\qquad otherwise
\end{array} \right.\end{equation}

Where ${\bm{w}_{ig}^v}$ denotes the $g$th neighbor between ${x_i^v}$ and ${x_j^v}$ in ${\bm{Q}(x_i^v)}$, i.e.${\bm{u}_g^v}$. In order to ensure the symmetry of matrix $S$, we need to operate with ${\bm{S}^v} = {{\left( {{{\left( {{\bm{S}^v}} \right)}^T} + {\bm{S}^v}} \right)} \mathord{\left/{\vphantom {{\left( {{{\left( {{\bm{S}^v}} \right)}^T} + {\bm{S}^v}} \right)} 2}} \right.\kern-\nulldelimiterspace} 2}$ . We consider setting weights for similar matrices from different views rather than simply accumulating similar matrices. i.e.$\bm{S} = \sum\nolimits_{v = 1}^k {{\bm{a}_v}} {\bm{S}^v}$, where ${\bm{a}_v}$ is a weight vector. Therefore, the similarity preservation part can be calculated as follows:


\begin{equation}\begin{array}{l}
\mathop {\min }\limits_B {\sum\limits_{i,j} {{\bm{S}_{ij}}\left\| {{\bm{b}_i} - {\bm{b}_j}} \right\|} ^2}\\
s.t.{\bm{B}} \in {\left\{ { - 1,1} \right\}^{r \times N}}{,\bm{B}}1 = 0,{\bm{B}}{{\bm{B}}^T} = n{\bm{I}_r} \label{blb}
\end{array}\end{equation}

Where ${\bm{b}_i} \in {\left\{ {- 1,1} \right\}^{r \times 1}}$ is the hash code representation to the $i$th sample, and $r$ is the length of the hash code. The last two constraints force the binary codes to be uncorrelated and balanced, respectively. Eq.(\ref{blb}) can be organized as:

\begin{equation}\begin{array}{l}
\mathop {\min }\limits_{\bm{B},\bm{W},\bm{a}} {\rm{ }}tr\left( {\bm{B}\bm{L}{\bm{B}^T}} \right)\\
s.t.{\rm{ }}\bm{B} \in {\left\{ { - 1,1} \right\}^{r \times N}},{\bm{B}}1 = 0,{\bm{B}}{{\bm{B}}^T} = n{\bm{I}_r}
\end{array}\end{equation}

Where ${\bm{L}=\bm{D}-\bm{S}}$, $\bm{D}$ is a diagonal matrix given by ${\bm{D}_{ii}} = \sum\nolimits_{j = 1}^N {{\bm{S}_{ij}}} $, and $\bm{L}$ is the graph Laplcian matrix.

\subsection{Overall Objective Function}
In order to learn binary codes associated with the clustering task, we find that the binary code representation learning and the discrete similarity preserving is both crucial.  At last, We combine similarity preservation with binary code learning into a common framework as follows:

\begin{equation}
\begin{array}{l}
\mathop {\min }\limits_{\bm{B},{\bm{H}^v},\bm{w},a} {\rm{ }}\sum\limits_{v = 1}^k {{{({a^v})}^c}} 

\left( {\underbrace{\left\| {\bm{B} - {\bm{H}^v}\bm{G}(\bm{x}_i^v)} \right\|_F^2{\rm{ + }}\delta \left\| {{\bm{H}^v}} \right\|_F^2}_{\text{binary code learning }}} \notag\right.  \\
\phantom{=\;\;}	
\left. -\frac{\lambda }{n}tr\underbrace{\left[ {{\bm{H}^v}\bm{G}(x_i^v){{\left( {{\bm{H}^v}\bm{G}(\bm{x}_i^v)} \right)}^T}} \right]}_{\text{balanced binary code partition }} \right) + \underbrace{\beta tr\left( {\bm{B}\bm{L}{\bm{B}^T}} \right)}_{\text{Graph based binary code }}\\
s.t.{\rm{  }}\sum\limits_v {{a^v}}  = 1,{a^v} > 0,\bm{B} \in {\left\{ { - 1,1} \right\}^{r \times n}},\bm{B}1 = 0,\bm{B}{\bm{B}^T} = n{\bm{I}_r} 
\label{total}
\end{array}
\end{equation}

where ${\delta}$, ${\beta}$ and ${\lambda}$ are regularization parameters to balance the effects of different terms. To optimize the complex discrete coding problems, an alternating optimization algorithm is proposed in the next section.

\section{Optimization Algorithm}

We have constructed a general framework named GMBL which can combine discrete hashing representation and structured binary clustering for multi-view data. We apply an alternative iterative optimization strategy to optimize the proposed objective function. The problem is resolved to separate the problem into several, which are to update a variable while fixing the remaining variables until convergence. In order to fully understand the proposed GMBL method, we summarize in Algorithm \ref{alg:Framwork}.

	
    \textbf{ Updating ${\bm{H}^v}$	}: When fixing other variables, we update the projection matrix by: 	
     \begin{equation}
	\min \Phi ({\bm{H}^v}){\rm{  = }}\left\| {\bm{B} - {\bm{H}^v}\bm{G}(\bm{x}_i^v)} \right\|_F^2{\rm{ + }}\delta \left\| {{\bm{H}^v}} \right\|_F^2 - \frac{\lambda }{n}tr\left[ {{\bm{H}^v}\bm{G}(\bm{x}_i^v){{\left( {{\bm{H}^v}\bm{G}(\bm{x}_i^v)} \right)}^T}} \right] \label{H}
     \end{equation}		
	
	 It closed-form solution can be obtained by setting partial derivative ${\frac{{\partial \Phi ({\bm{H}^v})}}{{\partial {\bm{H}^v}}} = 0}$, whose optimal solution is ${\bm{H}^v} = \bm{B}{\left( {\bm{G}({x^v})} \right)^T}{\bm{P}^{ - 1}}$, where 
	 $\bm{P} = (1 - \frac{\lambda }{n})\bm{G}({x^v}){\left( {\bm{G}({x^v})} \right)^T} \\+ \delta \bm{I}$.
		
   \textbf{Updating ${\bm{B}}$	}: We next move to update ${\bm{B}}$, the sub-problem with respect to the ${\bm{B}}$ are defined as follows:
	
     \begin{equation}	
	\begin{array}{l}
	\mathop {\min }\limits_B {\rm{ }}\Phi (\bm{B}) = \sum\limits_{v = 1}^k {{{({a^v})}^c}} \left\| {\bm{B} - {\bm{H}^v}\bm{G}(\bm{x}_i^v)} \right\|_F^2 + \beta tr\left( {\bm{B}\bm{L}{\bm{B}^T}} \right) + \mu {\left\| {\bm{B}{\bm{B}^T}} \right\|^2} + \rho {\left\| {\bm{B}1} \right\|^2}\\
	s.t.{\rm{  }}\bm{B} \in {\left\{ { - 1,1} \right\}^{r \times n}}  
	\end{array}	
     \end{equation}		
     	
     We design an effective algorithm that can maintain discrete constraints in the optimization process, and through this method we can obtain more efficient binary codes \cite{shen2015supervised},\cite{shen2016fast}. According to the DPLM algorithm, we can get ${\bm{B}}$ as follows:	
	
	\begin{equation}\nabla \Phi (\bm{B}) =  - 2tr\left[ {{\bm{B}^T}\sum\limits_{v = 1}^k {{{\left( {{a^v}} \right)}^r}{\bm{H}^v}\bm{G}\left( {{\bm{x}^v}} \right)} } \right] + 2\beta \bm{B}\bm{L} + \mu \bm{B}{\bm{B}^T}\bm{B} + \rho \bm{B}{11^T} \label{B}  \end{equation}
	
	Where $\nabla \Phi (\bm{B})$ is the gradient of $\Phi (\bm{B})$. We update variable $\bm{B}$ use to ${\bm{B}^{q + 1}} = {\mathop{\rm sgn}} ({\bm{B}^q} - \frac{1}{\eta }\nabla \Phi ({\bm{B}^q}))$ in each iteration.
	
  \textbf{Updating $a^v$}: According to the attributes of different views, Optimization of $a^v$ can be equivalent as the following optimization problem:
  
	\begin{equation}
	\begin{array}{l}
	\mathop {\min }\limits_{{a^v}}  \Phi \left( {{a^v}} \right) =\\ {\rm{ }}\sum\limits_{v = 1}^k {{{({a^v})}^c}} \left( {\left\| {\bm{B} - {\bm{H}^v}\bm{G}(\bm{x}_i^v)} \right\|_F^2{\rm{ + }}\delta \left\| {{\bm{H}^v}} \right\|_F^2 - \frac{\lambda }{n}tr\left[ {{\bm{H}^v}\bm{G}(x_i^v){{\left( {{\bm{H}^v}\bm{G}(\bm{x}_i^v)} \right)}^T}} \right]} \right)\\
	s.t.{\rm{  }}\sum\limits_v {{a^v}}  = 1,{a^v} > 0
	\end{array} \label{A}
    \end{equation}
    
    Let ${\mathcal{M} ^v} = \left\| {\bm{B} - {\bm{H}^v}\bm{G}(\bm{x}_i^v)} \right\|_F^2{\rm{ + }}\delta \left\| {{\bm{H}^v}} \right\|_F^2 - \frac{\lambda }{n}tr\left[ {{\bm{H}^v}\bm{G}(\bm{x}_i^v){{\left( {{\bm{H}^v}\bm{G}(\bm{x}_i^v)} \right)}^T}} \right]$ then we can rewritten (\ref{A}) as
	\begin{equation}
    \begin{array}{l}
    \mathop {\min \Phi \left( {{a^v}} \right) = }\limits_{{a^v}} \sum\limits_{v = 1}^k {{{({a^v})}^c}} {\mathcal{M} ^v}\\
    s.t.{\rm{  }}\sum\limits_v {{a^v}}  = 1,{a^v} > 0  
    \end{array}	 \label{AA}
    \end{equation}	
    
    We can solve the constraint equation by Lagrange multiplier method, the Lagrange function of (\ref{AA}) is  $\mathop {\min }\limits_{{a^v}} \Psi \left( {{a^v},\mathcal{M} } \right) = {\rm{ }}\sum\limits_{v = 1}^k {{{({a^v})}^c}} {\mathcal{M} ^v} - \mathcal{M} \left( {\sum\limits_v {{a^v}}  - 1} \right)$
    
    By setting the partial derivative of $\Psi \left( {{a^v},\mathcal{M} } \right)$ with respect to $a^v$ and $\mathcal{M}$ to zero, we can get  
    \begin{equation}
	\left\{ \begin{array}{l}
	\frac{{\partial \Psi }}{{\partial {a^v}}} = c{({a^v})^{c - 1}}{\mathcal{M} ^v} - \mathcal{M}  = 0,{\rm{ }}v = 1,2,...,k\\
	\frac{{\partial \Psi }}{{\partial \mathcal{M} }} = {\rm{         }}\sum\limits_{v = 1}^k {{a^v} - 1}  = 0
	\end{array} \right.
	\end{equation}
	Therefore, we can get $a^v$ as 	
	
	\begin{equation}
	{a^v} = \frac{{{{({\mathcal{M} ^v})}^{\frac{1}{{1 - c}}}}}}{{\sum\nolimits_v {{{({\mathcal{M} ^v})}^{\frac{1}{{1 - c}}}}} }}
    \end{equation}
	
	In order to obtain the local optimal solution, we update the three variables iteratively until the convergence.	
	

\begin{algorithm}[htb]
 	\caption{ Framework of ensemble learning for our method.}
 	\label{alg:Framwork}
 	\begin{algorithmic}[1]
 		\Require  
 		Data set $ \left\{ {\bm{X}^v} \right\}_{v=1}^{\rm{s}}$;
 		Anchor samples $\left\{  {a_j^v} \right\} _{j=1}^{\rm{s}}$;
 		parameters $\beta$, $\lambda$ and $\delta$; 	
 		\Ensure 
 		binary code $\bm{B}$;
 		\State Initialize binary code $\bm{B}$ ; nonlinear embedding ${\bm{G}{(\bm{x}_i^v)}}$ ;  Weights from different views $\left\{ {a^v} \right\} _{v=1}^{\rm{k}} = \frac{1}{k}$; project matrix ${\bm{H}^v}$; binary code length $r$;
 		\label{code:fram:extract}
 		\State Construct $\bm{S}$ get laplacian matrix ;
 		\label{code:fram:trainbase}	
 		\Repeat;	
 		\State Update ${\bm{H}^v}$  by solving equation in Eq.\ref{H};  
 		\label{code:fram:add}
 		\State Update binary code $\bm{B}$ according to Eq.\ref{B};
 		\label{code:fram:classify}
 		\State Update ${a^v}$ by solving problem Eq.\ref{A};
 		\label{code:fram:select} 
 		\Until convergence ;     
 	\end{algorithmic}
\end{algorithm}

\section{Experimental Evaluation}

In this section, extensive experiments are the command to evaluate the proposed binary clustering methods in clustering performance. All the experiments are conducted with Matlab 2018b using a standard Windows PC with an Intel 3.4 GHz CPU.

\subsection{ Experimental Settings }
In this section, we describe the datasets and comparison methods. We evaluated the clustering performance of GMBL by comparing it with several classical hash methods in the multi-view datasets. In addition, the effectiveness of GMBL algorithm is evaluated by comparing the real-valued multi-view methods. In the end, we compared the single-view low-dimensional embedding in the framework with the original GMBL low-dimensional embedding to verify that our method can modify and supplement complementary information between different views.

\subsubsection{Datasets}
In most practical applications, images are generally represented by multiple features, which constitute multiple views of the experiment. Without loss of generality, we evaluated our image cluster method using five popular datasets. Some Image samples of datasets are presented in Fig.\ref{Fig.2}. The details of utilizing the data information are listed as follows:

\begin{figure}[htb]
	\centering	
	\subfigure[Some images from Caltech256. It consists of 101 classes in total.]{	
		\label{fig:a ACC} 
		\includegraphics[width=9.8cm]{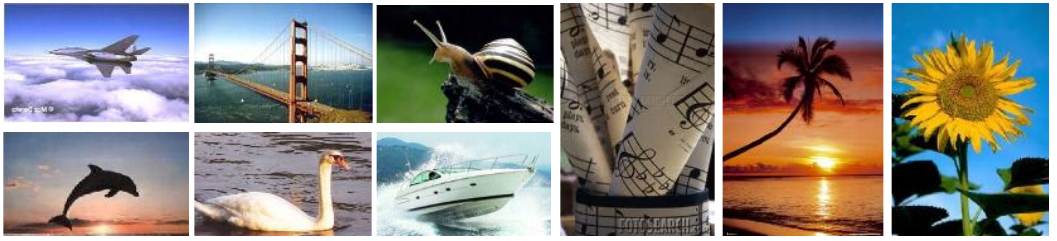}}
	\subfigure[Some images from Caltech101. Caltech101 is an image dataset which contains 101 classes and 1 backgroud	class.]{
		\label{fig:subfig:b} 
		\includegraphics[width=9.8cm]{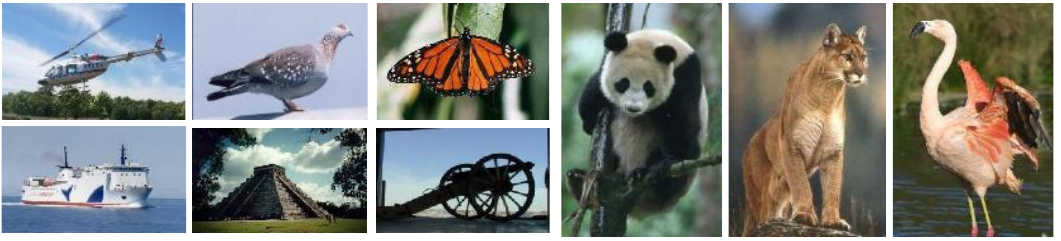}}
	\subfigure[Some images from Coil-100. There are 100 classes in this dataset. ]{
		\label{fig:subfig:c} 
		\includegraphics[width=9.8cm]{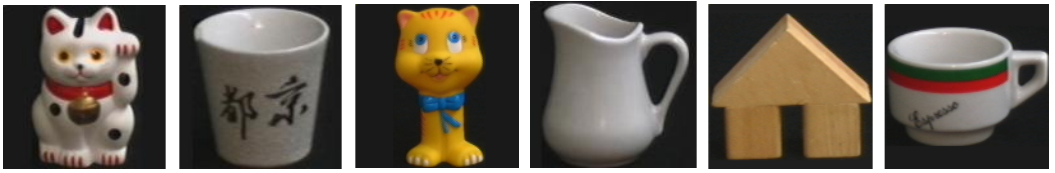}}
	\caption{Some sample images of these image datasets for various applications.}	
	\label{Fig.2} 
\end{figure}

\textbf{Caltech101}\footnotemark[1]\footnotetext[1]{http://www.vision.caltech.edu/ImageDatasets/Caltech101/} contains 9144 images associated with 101 objects and a background category. It is a benchmark image dataset for image clustering and retrieval tasks. Each example is associated with a reciprocally class label. For this dataset, five publicly available features are engaged for experiments, i.e. 48-dim Gabor feature, 928-dim LBP feature, 512-dim GIST feature, 254-dim CENTRIST feature, 40-dim wavelet moments and 1984-dim HOG feature.

\textbf{Caltech256}\footnotemark[2]\footnotetext[2]{http://www.vision.caltech.edu/Image-Datasets/Caltech256/} contains 30,607 images of 256 object categories, each of which contains more than 80 images. We use a 729-dim color histogram feature, 1024-dim Gist feature and 1152-dim HOG feature, which three different types of features.

\textbf{NUS-WIDE-obj}\footnotemark[3]\footnotetext[3]{http://lms.comp.nus.edu.sg/research/NUS-WIDE.htm} contains 30,000 images in 31 categories. The features of the dataset can be found on the contributor's home page, including 65-dim color histogram(CH), 226-dim color moments(CM),74-dim edge distribution(ED), 129-dim wavelet texture(WT) and 145-dim color correlation(CORR).

\textbf{Coil-100}\footnotemark[4]\footnotetext[4]{http://www1.cs.columbia.edu/CAVE/software/softlib/coil-100.php} is the abbreviation of the Columbia object image library dataset, which consists of 7200 images in 100 object categories. Each category contains 72 images and all images are with size 32×32. Intensity, 220-dim DSD, 512-dim HOG and 768-dim Gist features are extracted for representation.  

\textbf{CiteSeer}\footnotemark[5]\footnotetext[5]{http://lig-membres.imag.fr/grimal/data.html} consists of 3,312 documents on scientific publications. These documents can be further classified into six categories: Agent, AI, DB, IR, ML and HCI. For our multi-view learning clustering, we construct a 3703-dimensional vector representing the keywords of text view and a 3279-dimensional vector representing the reference relationship between another document. All the features of discretion are briefly summarized in table \ref{table 3}. 
%


\begin{table}[]
	\scriptsize	
	\centering	
	\caption{Summarization of each dataset}   
	\label{table 3}
	\begin{tabular}{cccccc} 
		\hline 
		Datasets &Caltech101 &Caltech256 &NUS-WIDE-obj &Coil-100  &	CiteSeer \\  \hline
		 Samples & 9144      & 30608     & 30000       & 7200     & 3312         \\
	    Classes  & 102       & 175       & 31          & 100      &  6       \\
		Views    & 6         & 3         & 5           & 3        & 2   \\	       \hline				
	\end{tabular}
\end{table}

\subsubsection{Compared Methods}

We compared our approach with the following state-of-the-art methods, including hash multi-view and real-value multi-view methods for clustering. As for the hash method, we utilized seven famous single-view hash algorithms and two multi-view hash clustering algorithms as comparing methods, including LSH \cite{gionis1999similarity}, MFH \cite{song2013effective}, SH \cite{weiss2009spectral}, DSH \cite{jin2013density}, SP \cite{xia2015sparse}, SGH \cite{jiang2015scalable}, BPH, ITQ \cite{gong2012iterative}, BMVC \cite{zhang2018binary}, HSIC \cite{zhang2018highly}. For single-view hash methods, we adopted the best result of each feature clustering.
As for the real-value multi-view method, we adopted seven algorithms as comparing methods, including k-means \cite{jetsadalak2018algorithm}, SC \cite{von2007tutorial}, Co-regularize \cite{kumar2011co}, AMGL \cite{nie2016parameter}, Mul-NMF \cite{liu2013multi}, MLAN \cite{nie2017multi}. It is noteworthy that the k-means method concatenates multi-view data into one vector as the evaluation result. The length of the hash code used in the experiment is 128-bits. We use the source code from the author's homepage for comparative experiments.

\subsubsection{Evaluation Metrics}

To generally evaluate the performance for clustering, We report the experimental results using four most widely used evaluation metrics, including accuracy(ACC), normalized mutual information(NMI), Purity and F-score \cite{gao2015multi}, \cite{cao2015diversity}. For all algorithms, the higher value of metrics indicates better performance. For the hashing methods, five different bits coding length are used for all datasets.

\subsection{Hash Method Experimental Results and Analysis}

In the section, we conducted experiments for hash clustering on 5 datasets (Include Caltech101, Caltech256, NUS-WIDE-obj, Coil-100 and CitySeer) to prove the performance of our proposed method.
We utilized all methods to project multi-view features into five hash codes of different lengths and adopted the k-means method to finish the task of image clustering. The results with different code lengths on four benchmark image datasets are reported in Figures \ref{Fig.4}, \ref{Fig.5}, \ref{Fig.6} and \ref{Fig.7}. Table \ref{table 4} shows the results when the hash code length is 128-bits in text clustering from two views. We have the following observations:
\begin{figure}	
	\centering	
	\subfigure[ACC]{	
		\label{fig:a ACC} 
		\includegraphics[width=3.8cm]{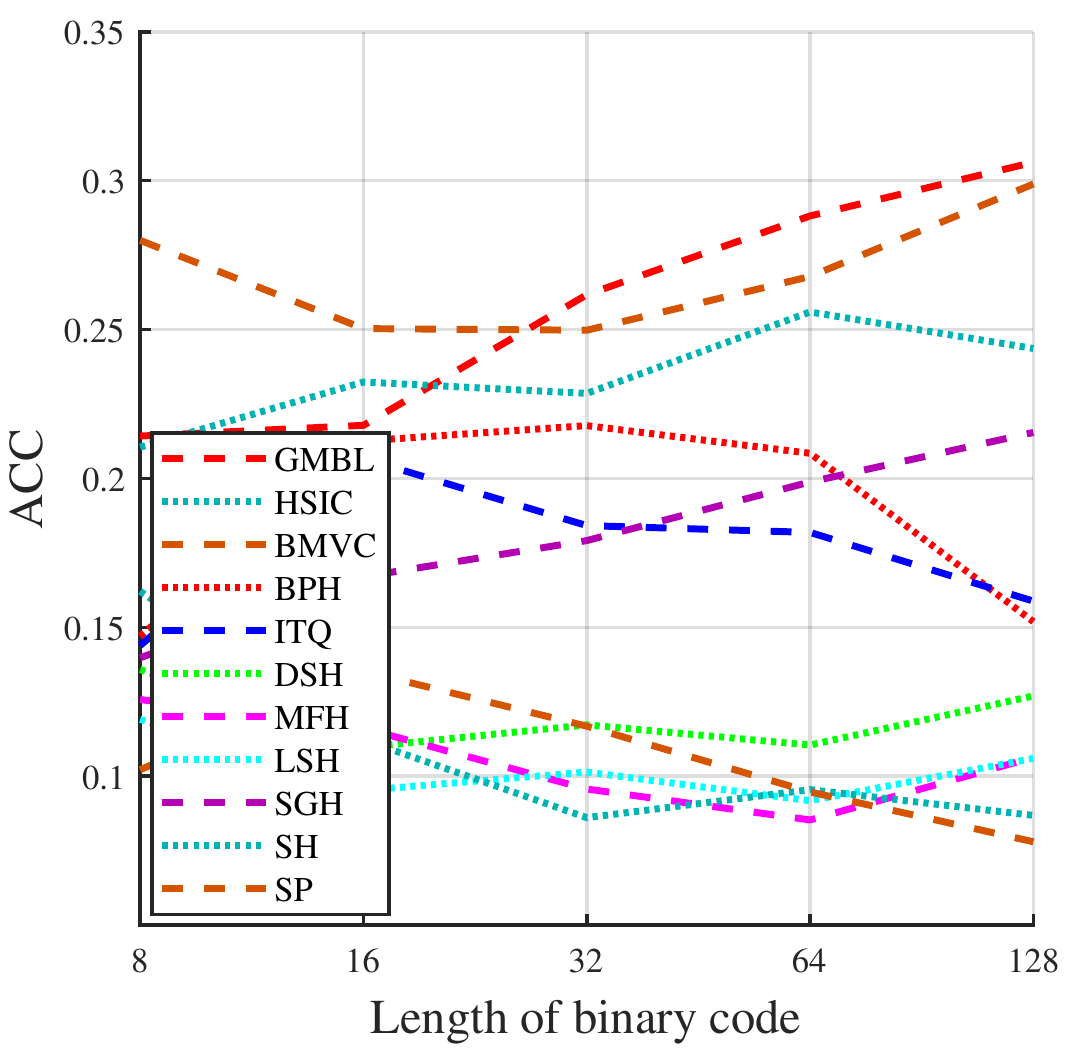}}
	\subfigure[NMI]{
		\label{fig:subfig:b} 
		\includegraphics[width=3.8cm]{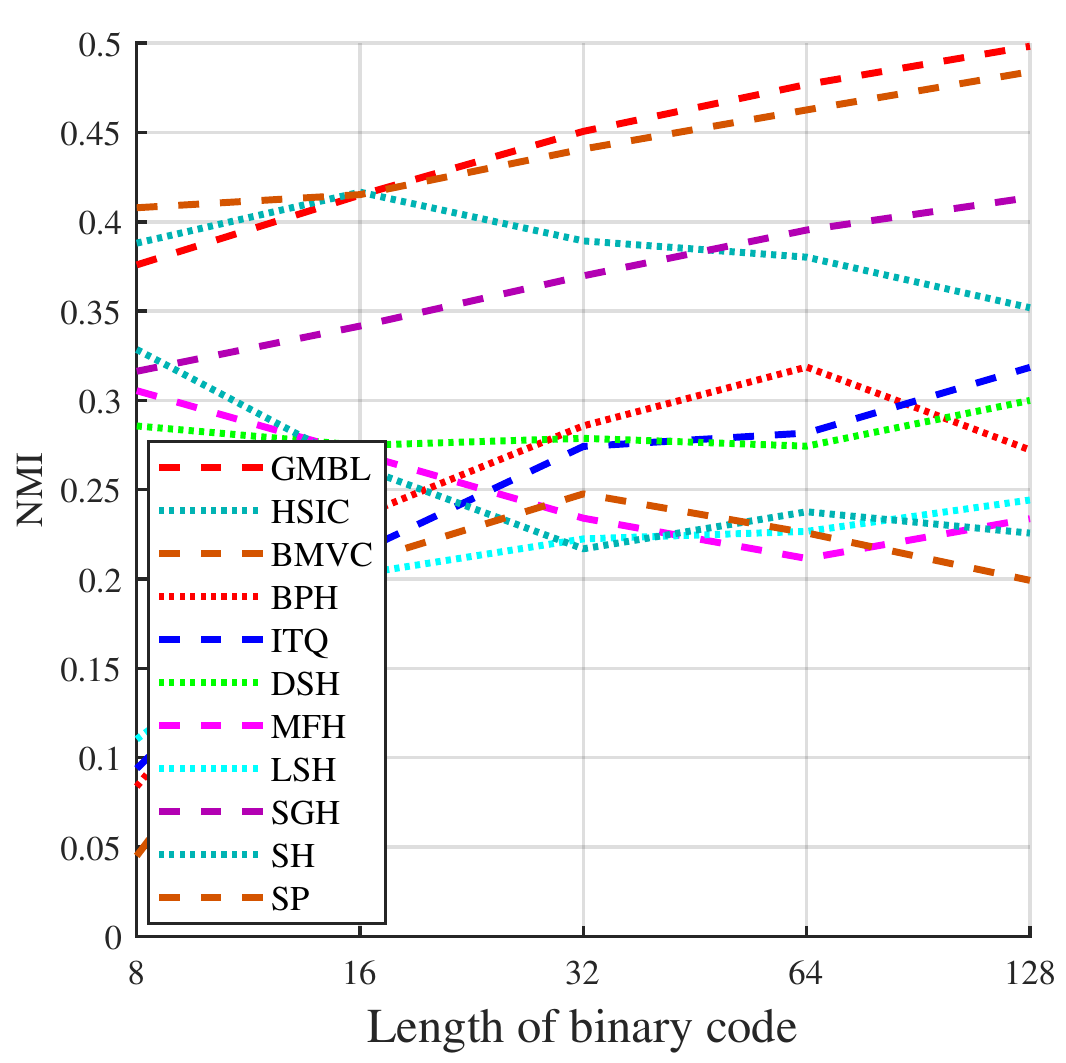}}
	\subfigure[Purity]{
		\label{fig:subfig:c} 
		\includegraphics[width=3.8cm]{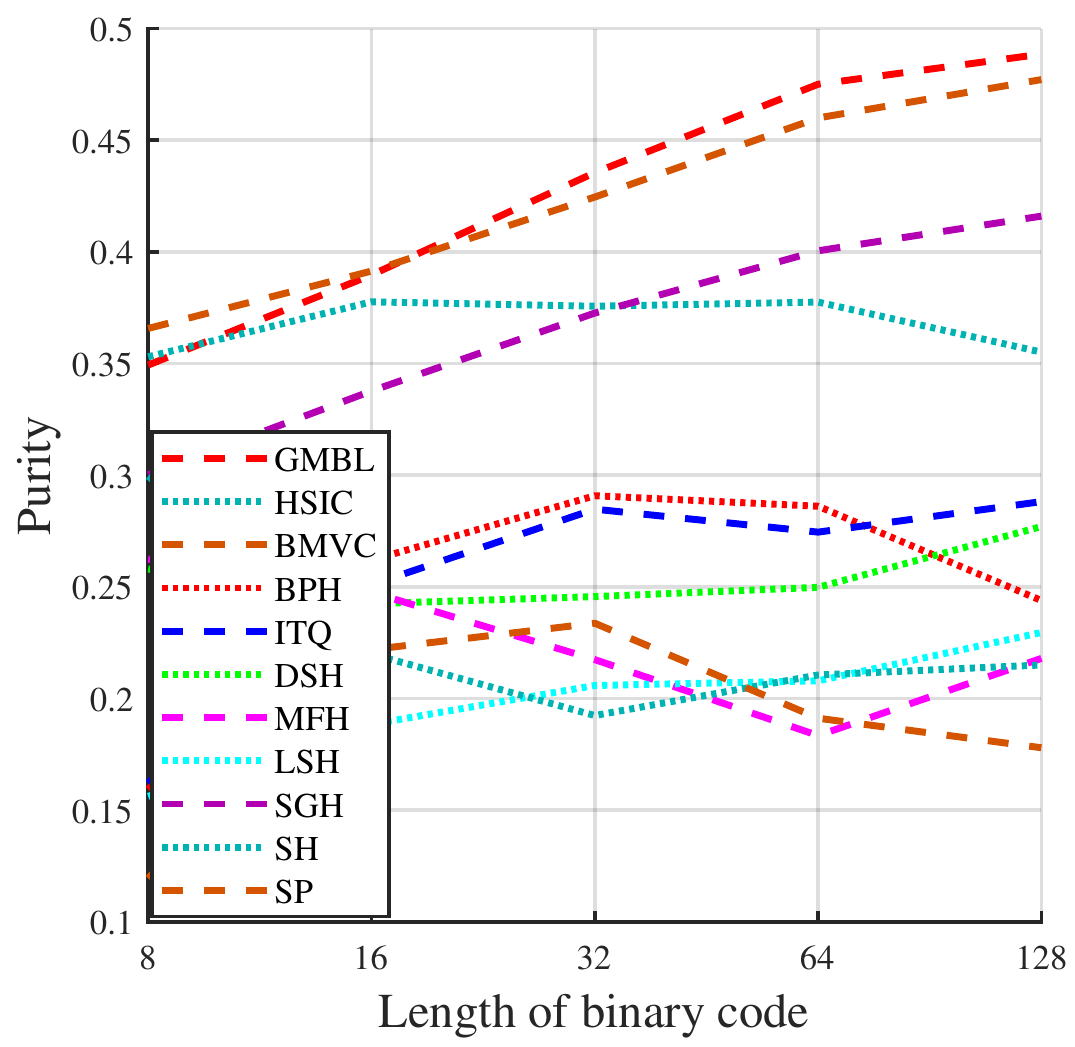}}
	\caption{
		Experiment results on caltech101. It is clear that our proposed GMBL can achieve the best performance in most situations.}	
	\label{Fig.4} 
\end{figure}

For Caltech101 datasets, we adopted six views to complete the clustering task, which is the dataset with the most views in the experiment. We adopt a view with the best clustering performance is used to evaluate single-view hash methods experiments. It is clear that GMBL can achieve better performance than the other hash methods in different binary code lengths. Generally, the results of multi-view algorithms are better than single-view hash ones. It shows that in GMBL, the result increases with the increase of the hash code length. GMBL can obtain better results compared with the multi-view hash methods. Because GMBL can construct a similarity matrix to obtain the nearest neighbor relation of data, the optimal result can be obtained when the length of the hash code increases. It can be found from Fig.\ref{Fig.4} that when the hash code length of our method is short, the clustering result can't obtain better performance. The reason may be the hash code length is short, which the nearest neighbor relationship of the data is not well preserved.
In the algorithm of this paper, several parameters have a significant influence on the experiment in Eq.(\ref{total}). Generally speaking, larger the values of $\delta $ and $ \frac{\lambda }{n} $, the experimental results attempt to lower. It is possible that the restriction of large regular terms will restrict the efficient learning of hash codes. Through many experiments found that increasing the value of $\beta$ will reduce the clustering performance while increasing the value of the nearest neighbor parameter $g$ will improve the clustering result of GMBL. However, the running time will be affected by the number of nearest neighbors and our method selects $6-9$ nearest neighbors for clustering. If more strategies are adopted to select anchor points, clustering performance will be improved. Besides, when the number of anchor point $s$ is greater than 1000, the clustering performance is not significantly improved.

\begin{figure}	
	\centering	
	\subfigure[ACC]{	
		\label{a} 
		\includegraphics[width=3.8cm]{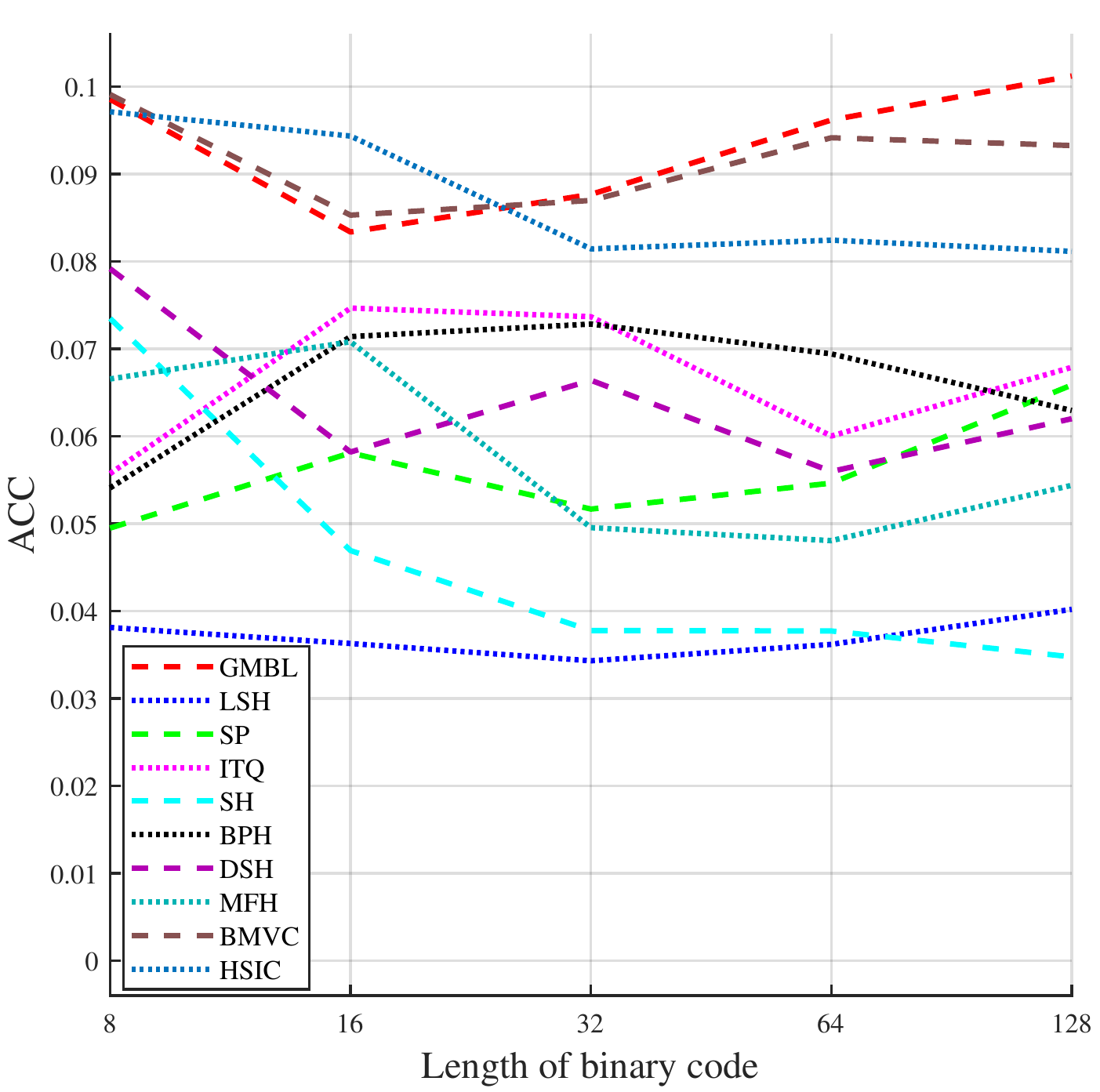}}
	\subfigure[NMI]{
		\label{b} 
		\includegraphics[width=3.8cm]{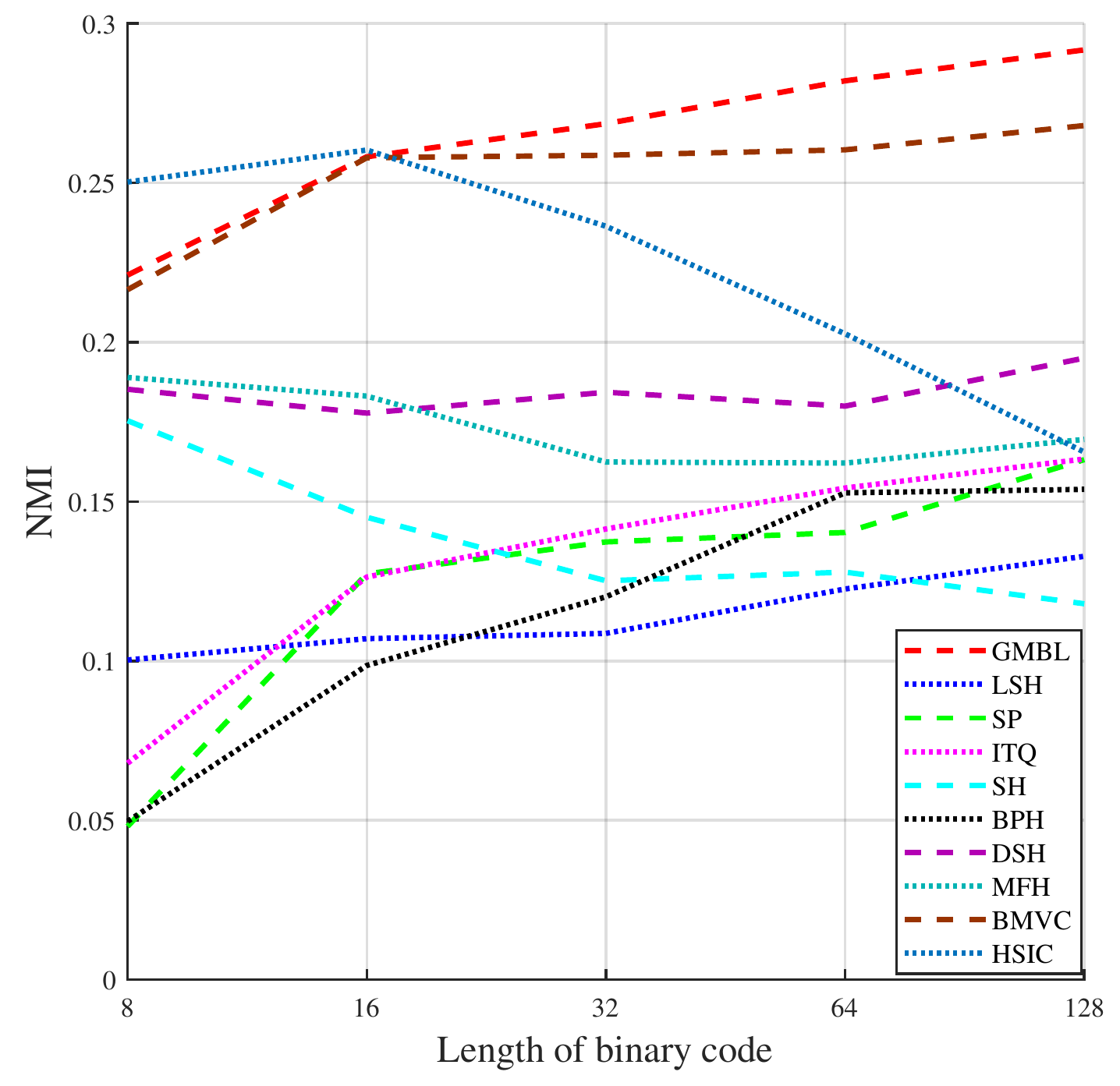}}
	\subfigure[Purity]{
		\label{c} 
		\includegraphics[width=3.8cm]{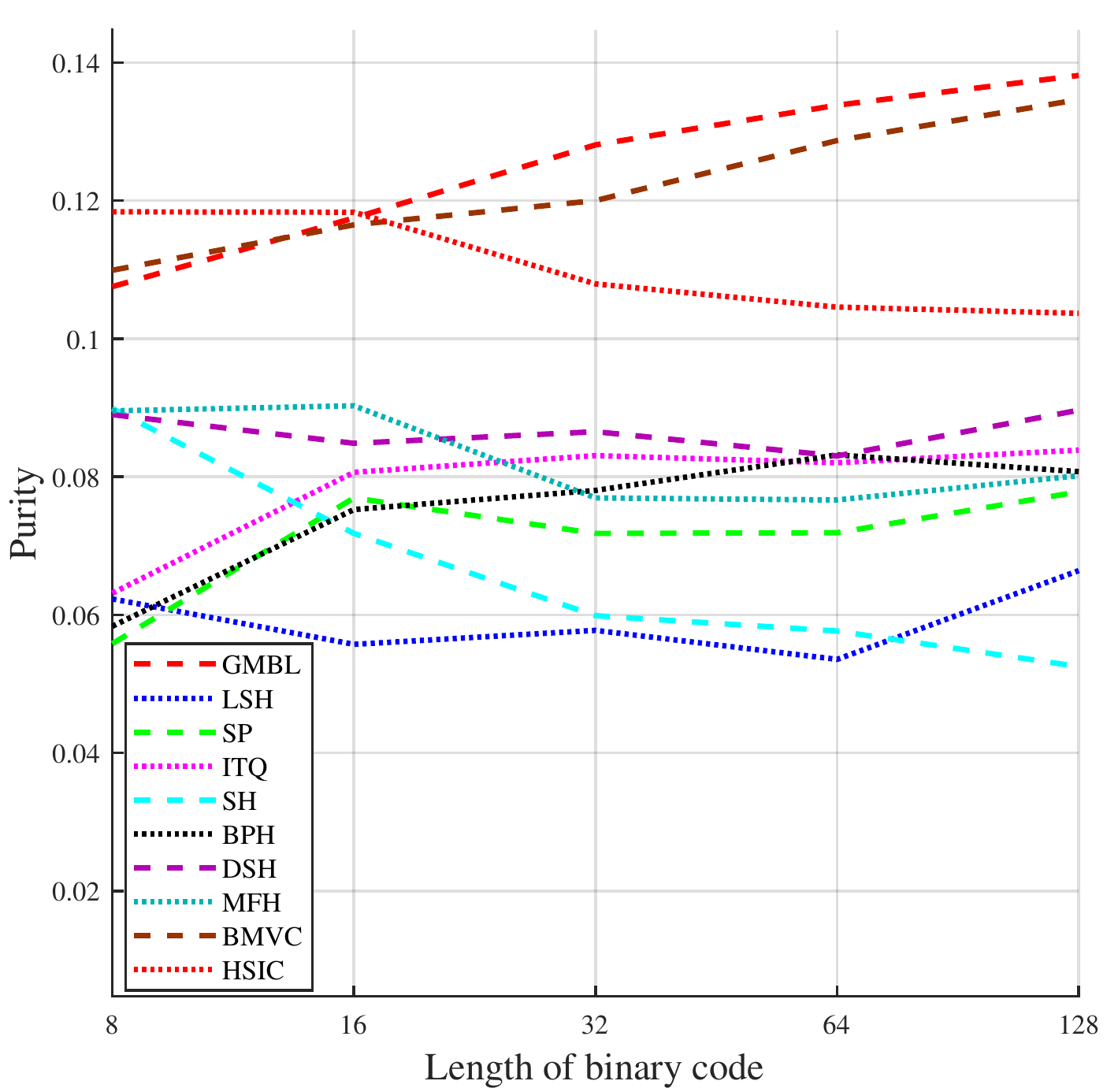}}
	\caption{Performance of different clustering methods vs. different code lengths of clusters on Caltech256.}	
	\label{Fig.5} 
\end{figure}

For the Caltech256 dataset, we randomly selected 175 categories of images as the experimental data with a total of 20222 images. It is explicit from Fig.\ref{Fig.5} that our approach obtains the perfect results in terms of ACC, NMI, and Purity among all the compared other hash methods in 128-bits binary code. From Fig.\ref{Fig.5}, we can observe that GMBL outperforms other hash methods when the code length is relatively large (i.e., greater than 32). As the length of the binary code increases, the performance of all algorithms improves. On the Caltech256 dataset, our model achieved 0.29 on NMI when the code length was 128-bits while the second-highest NMI was 0.26.

\begin{figure}[htb]	
	\centering	
	\subfigure[ACC]{	
		\label{a} 
		\includegraphics[width=3.8cm]{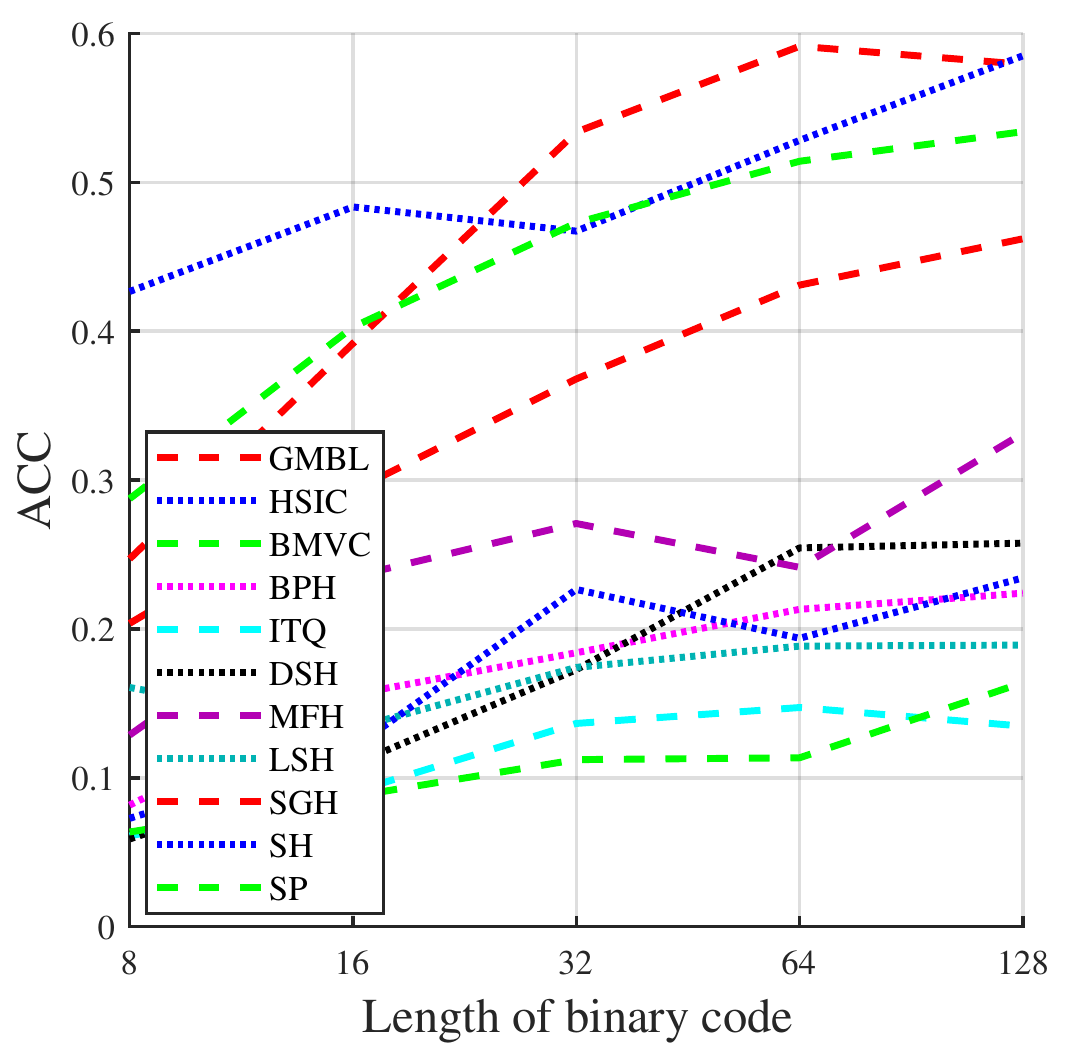}}
	\subfigure[NMI]{
		\label{b} 
		\includegraphics[width=3.8cm]{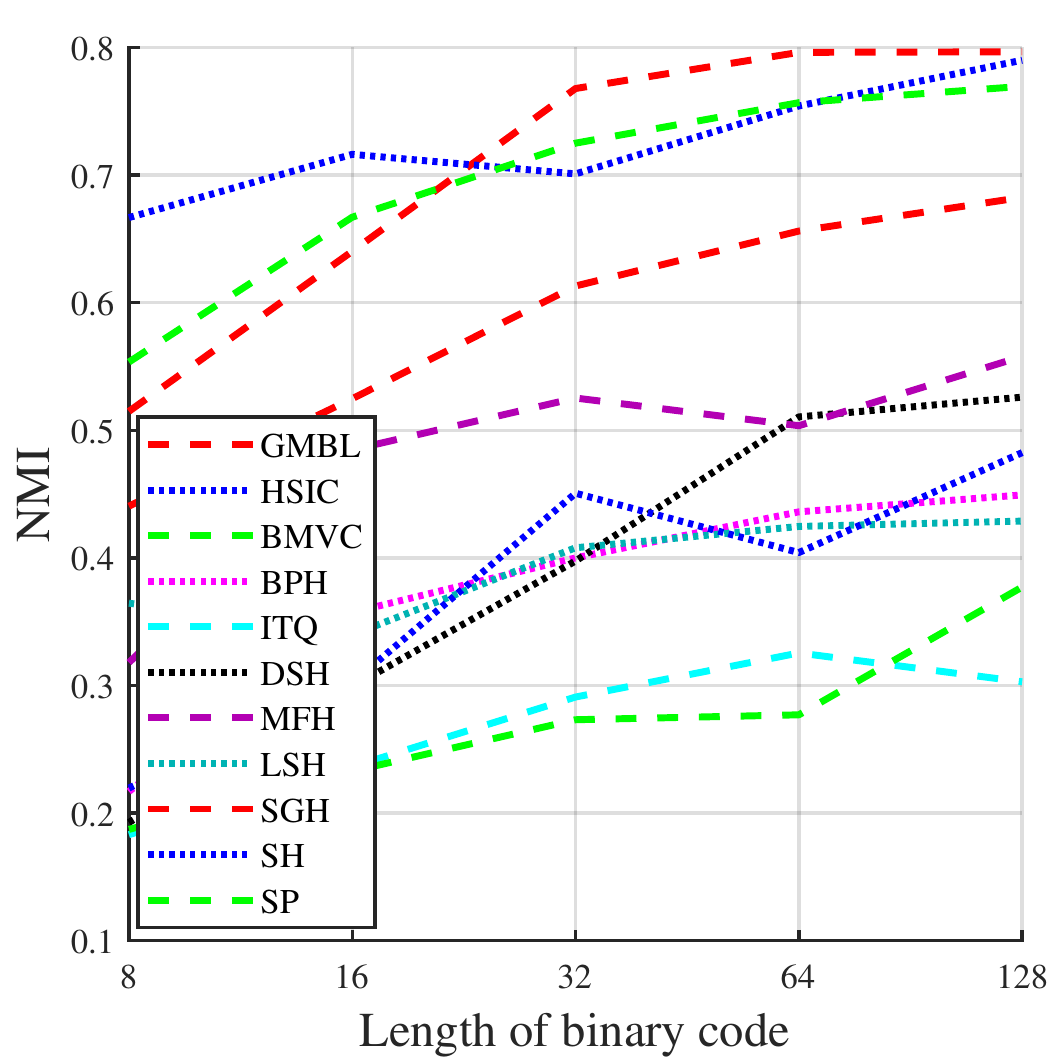}}
	\subfigure[Purity]{
		\label{c} 
		\includegraphics[width=3.8cm]{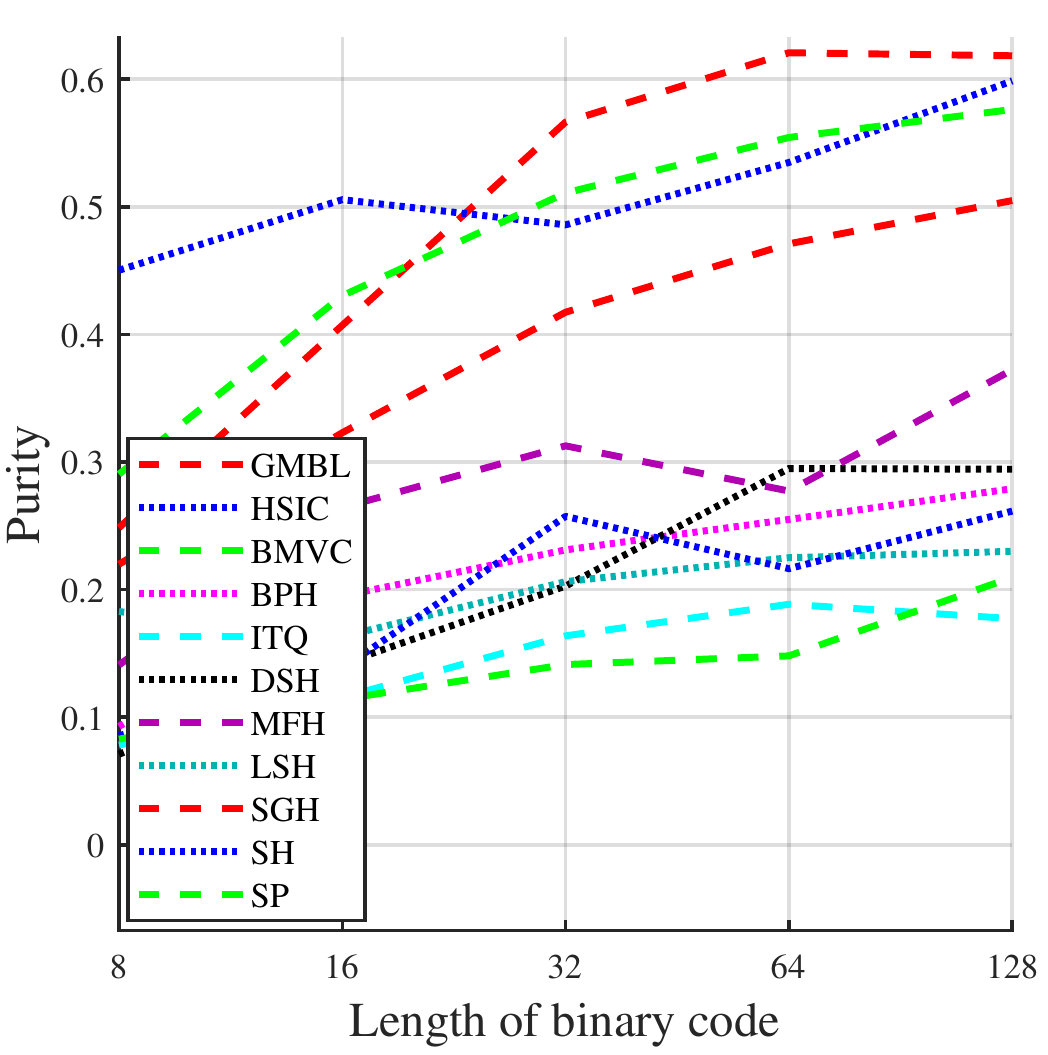}}
	\caption{ACC, NMI and Purity results on Coil-100.}	
	\label{Fig.6} 
\end{figure}

Fig.\ref{Fig.6} illustrates the experimental results on the Coil-100 dataset. Our method outperforms all the other methods of NMI and Purity evaluation metrics. For the ACC, HSIC method obtains the best result and we can deduce that using individual information and shared information to capture the hidden correlations of multiple views is necessary. We can also find that BMVC and HSIC can achieve the best performance in the short hash code lengthIt can be observed from the Fig.\ref{Fig.6} that the results of roughly all single-view hash methods are significantly poorer than that of multi-view hash methods under different hash code lengths.

\begin{figure}[htb]	
	\centering	
	\subfigure[ACC]{	
		\label{fig:a} 
		\includegraphics[width=3.85cm]{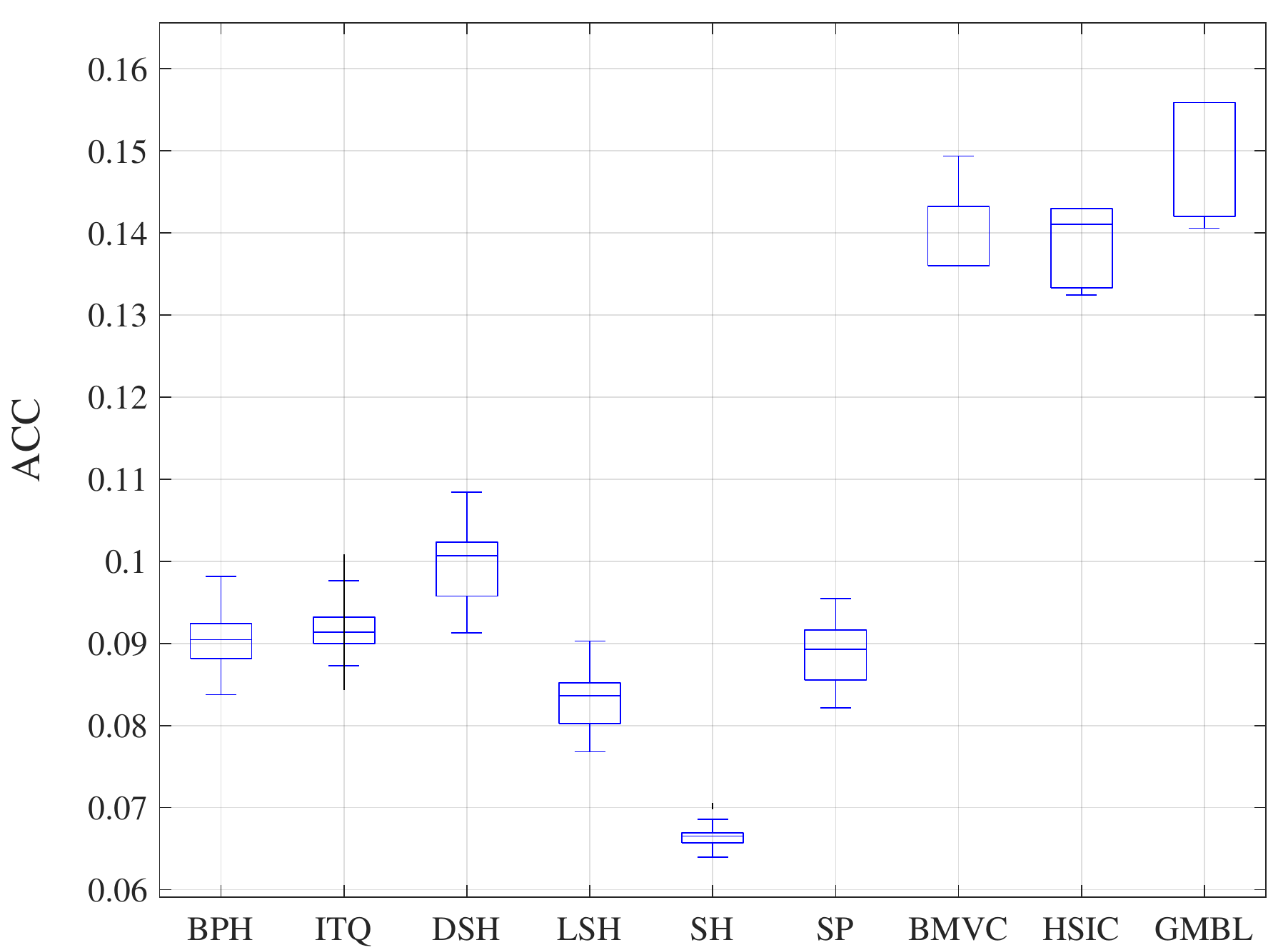}}
	\subfigure[NMI]{
		\label{fig:subfig:b} 
		\includegraphics[width=3.85cm]{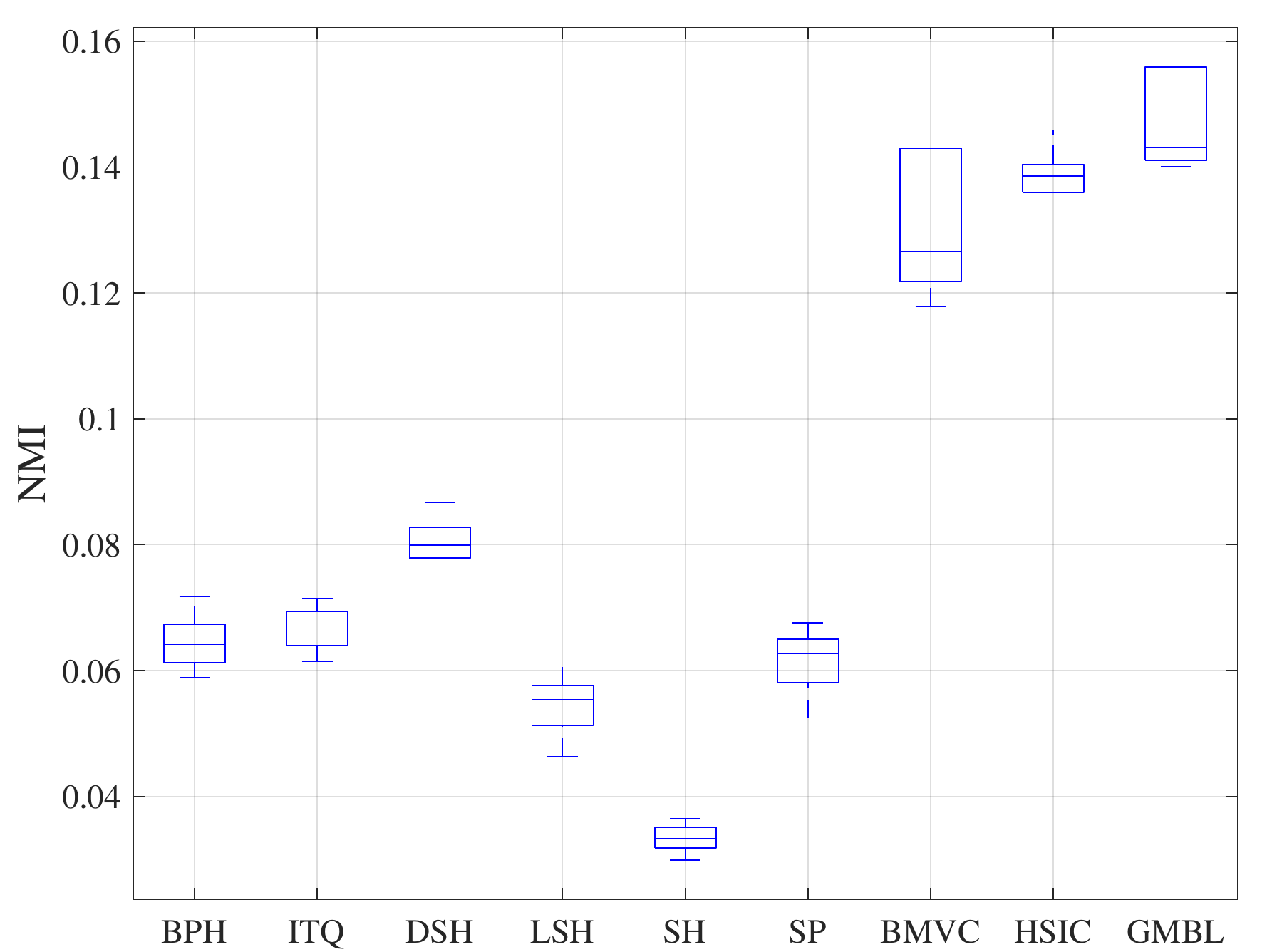}}
	\subfigure[Purity]{
		\label{fig:subfig:c} 
		\includegraphics[width=3.85cm]{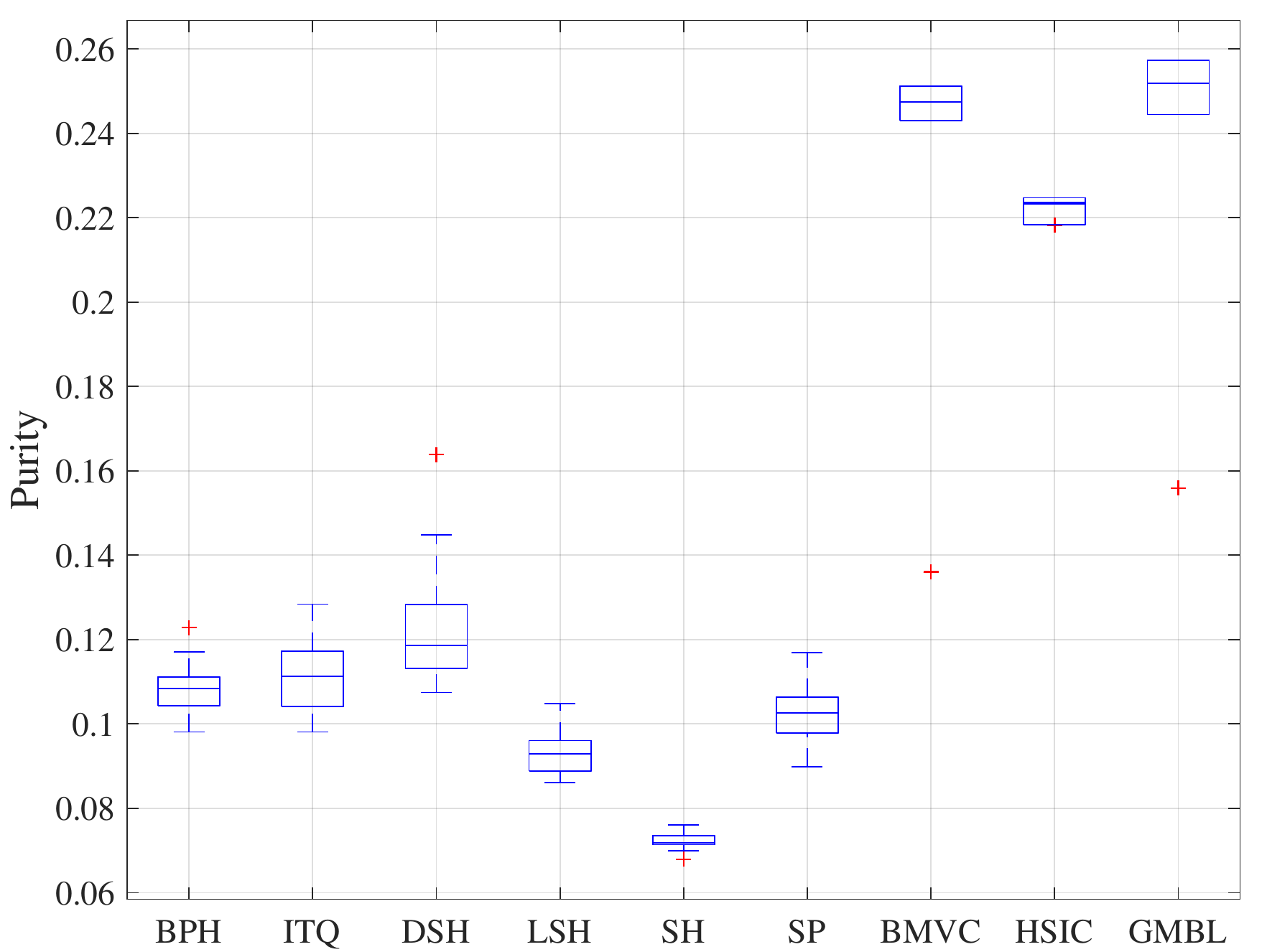}}
	\caption{ACC, NMI and Purity results on NUS-WIDE-obj. Multi-view methods achieve ideal performance.}	
	\label{Fig.7} 
\end{figure}

As we all know, NUS-WIDE-obj is a widely used dataset that includes 30000 images from Flickr. The dataset consists of 31 categories and each image is marked by at least one label and multiple labels can be assigned to each image. The dataset was divided into 12072 test sets and 17928 train sets by the provider. In this algorithm, we use the test set to evaluate the clustering task. For multiple labels with the same sample, we will automatically specify the corresponding labels ground-truth after the clustering task of each algorithm is completed. In Fig.\ref{Fig.7}, the ACC, NMI, and Purity of all hash algorithms under 128-bits binary code are reported. Box diagram corroborates the advantages of our GMBL relative to its simulated alternatives. 

%

Experiments demonstrate that the single-view hash method can also obtain satisfactory performance. Moreover, both SGH and ITQ get excellent performance. In general, multi-view methods get better results than single-view methods. Generally speaking, Multi-view methods exploit multi-view information and achieve better results. For multi-view data that holds more of the original data structures during the construction of the binary code that our method obtains a similarity matrix reflecting the local structure of the original data. Therefore, graph-based similarity matrix construction plays an important role in our method, which can effectively obtain the structural relationship between the initial input and adjacent data. GMBL has been verified can improve Largely clustering performance. We can notice that GMBL achieves much better results than BMVC on almost all datasets. The primary reason is that in the process of binary code, we can keep the local structure of data to further explore the internal relationship of data, so as to obtain better clustering results.

\begin{table}[]
	\tiny	
	\centering	
	\caption{The clustering results on CitySeer dataset}   
	\label{table 4}
	\begin{tabular}{p{0.8cm}|cccccccccc} 
		\hline 
    	Methods& LSH & SP & MFH    & ITQ    & SH     & BPH    & DSH    & BMVC   & HSIC& GMBL\\    \hline
		ACC    & 0.1413 & 0.1887 & 0.1954 & 0.2017 & 0.2126 & 0.1730 & 0.2292 & 0.2343 & 0.2560 & \textbf{0.2766} \\ 
		NMI    & 0.0105 & 0.0432 & 0.0079 & 0.0431 & 0.0026 & 0.0124 & 0.0353 & 0.0298 & 0.0298 & \textbf{0.0517} \\ 
		Purity & 0.2542 & 0.3222 & 0.2225 & 0.3206 & 0.2183 & 0.2292 & 0.2497 & 0.2844 & 0.2844 & \textbf{0.3273} \\ \hline
	\end{tabular}
\end{table}

To demonstrate the robustness of the GMBL, we consider the clustering experiment in the text dataset. There are 3312 documents in CitySeer dataset, which are divided into six categories. We use keywords and references between documents as two views for clustering experiments. Table \ref{table 4} compare ACC, NMI, and Purity when the length of the hash code is 128-bits. We have the following observations: GMBL obtained a higher value in the three indexes of the clustering task, which consistently outperforms other methods by large margins in all situations. Compared with single-view hashing method, ACC, NMI and Purity were increased 24\%-49\%, 17\%-90\% and 3\%-33\%; Compared with multi-view hashing method, ACC, NMI and Purity were improved 7\%-16\%, 42\%-60\% and 13\%-17\%; The results demonstrate that the proposed multi-view algorithm is effective in utilizing the graph-based method. 

\begin{figure}	
	\centering	
	\subfigure[]{	
		\label{fig:a} 
		\includegraphics[width=3.8cm]{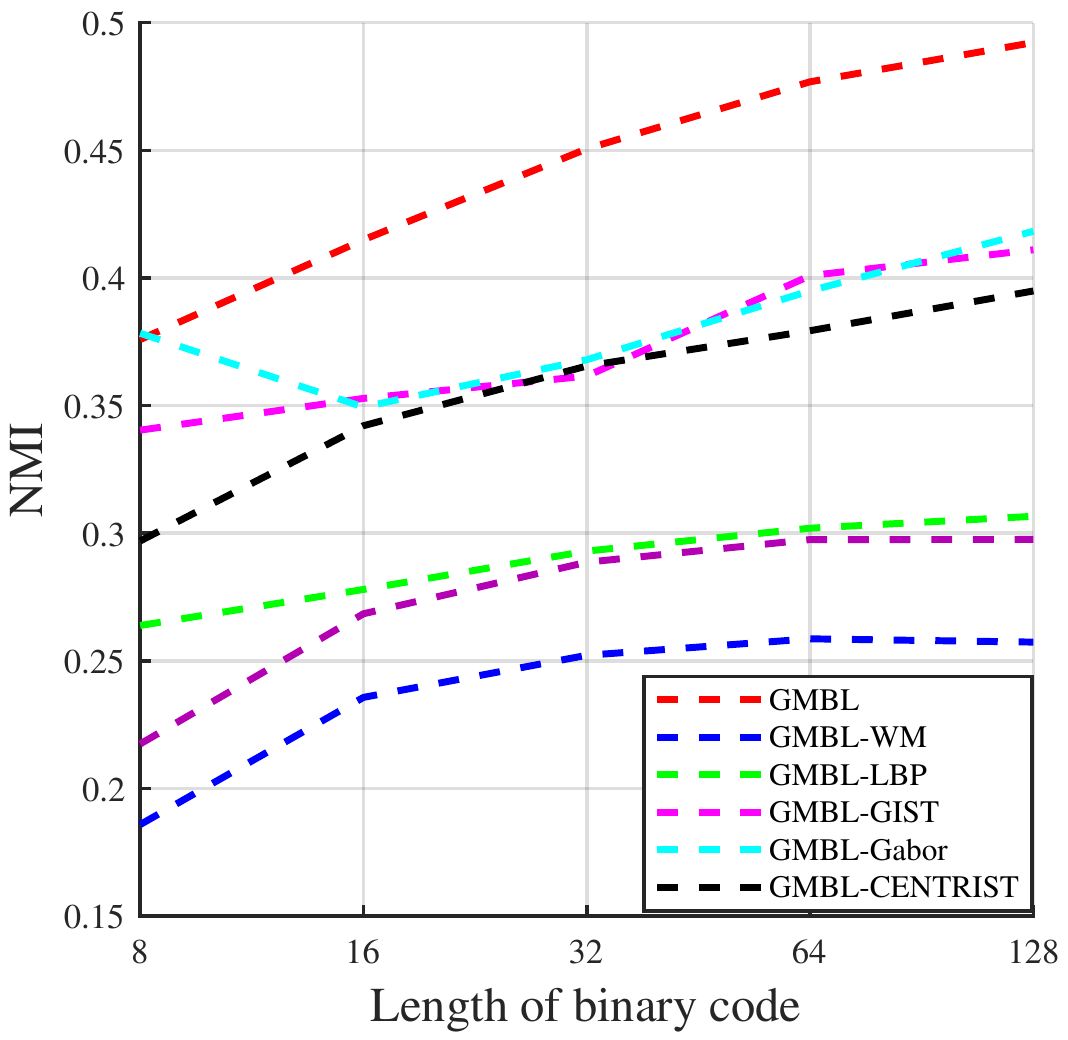}}
	\subfigure[]{
		\label{fig:subfig:b} 
		\includegraphics[width=3.8cm]{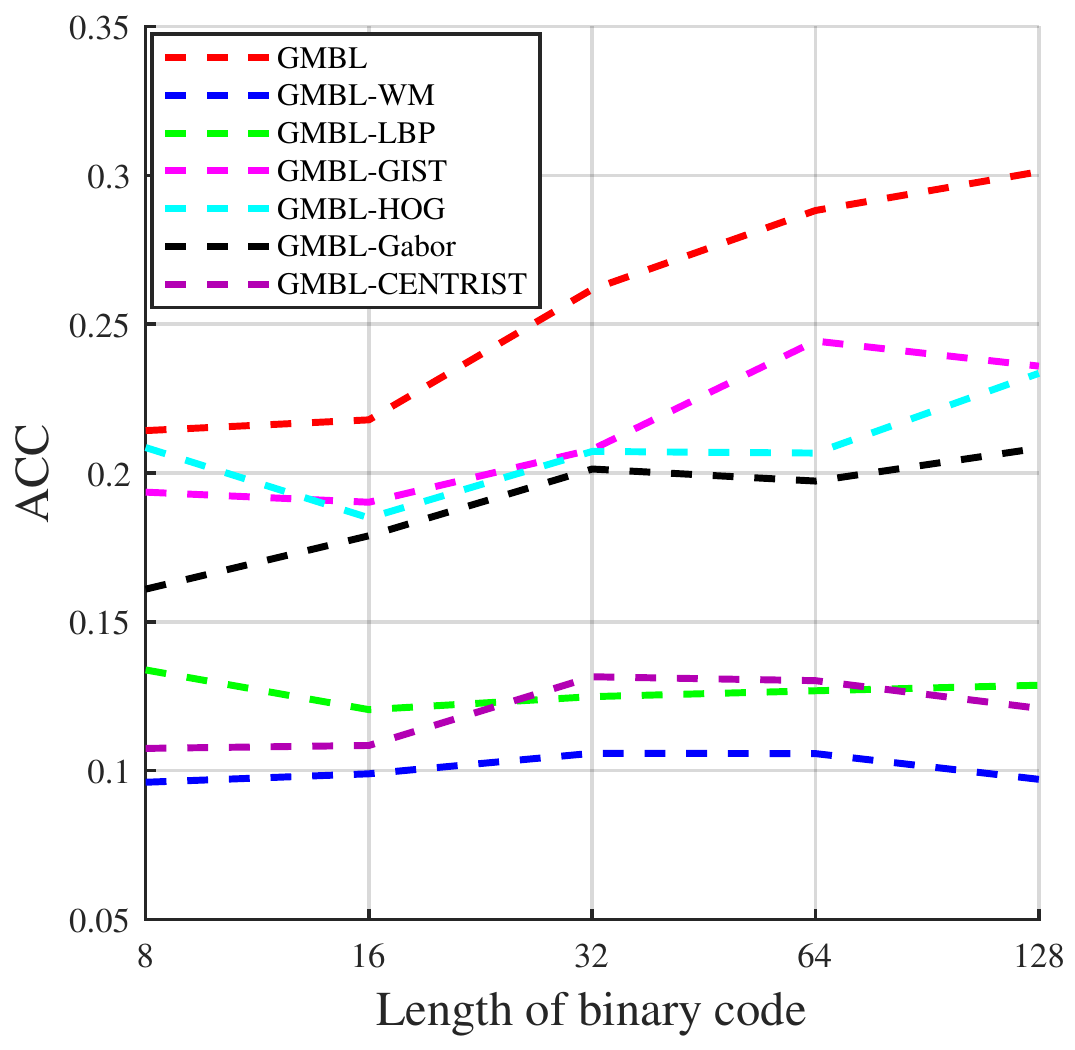}}
	\subfigure[]{
		\label{fig:subfig:c} 
		\includegraphics[width=3.8cm]{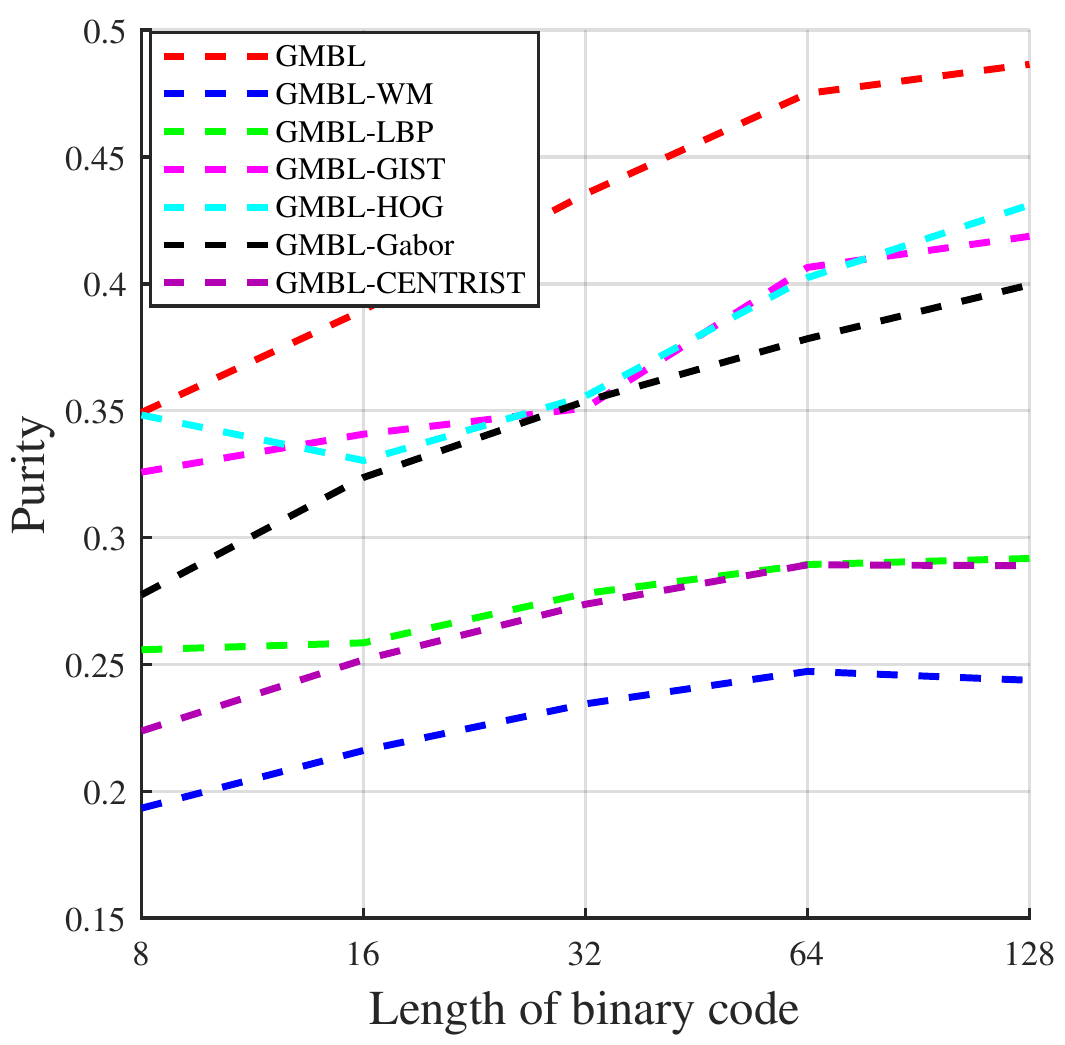}}
	\subfigure[]{
		\label{fig:subfig:b} 
		\includegraphics[width=3.8cm]{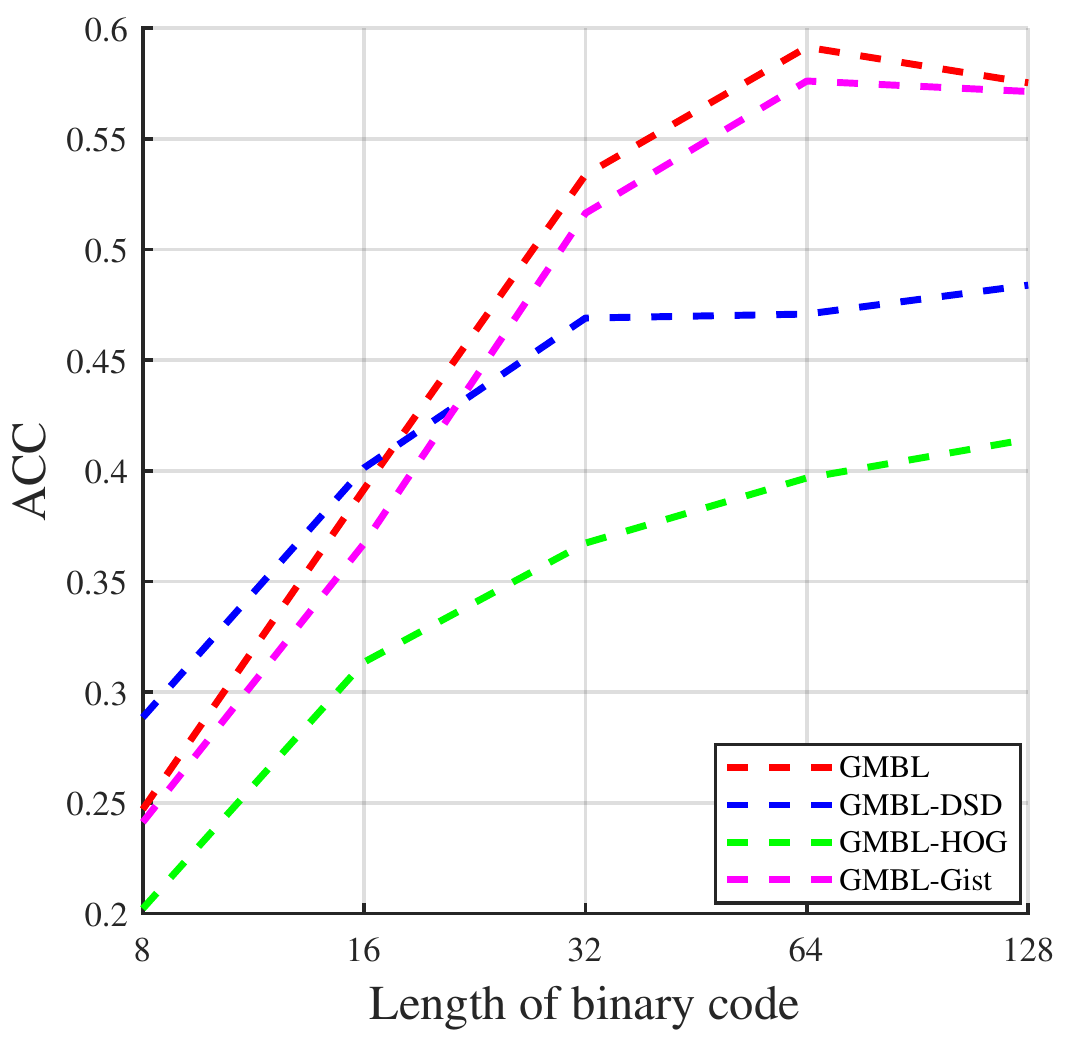}}
	\subfigure[]{
		\label{fig:subfig:b} 
		\includegraphics[width=3.8cm]{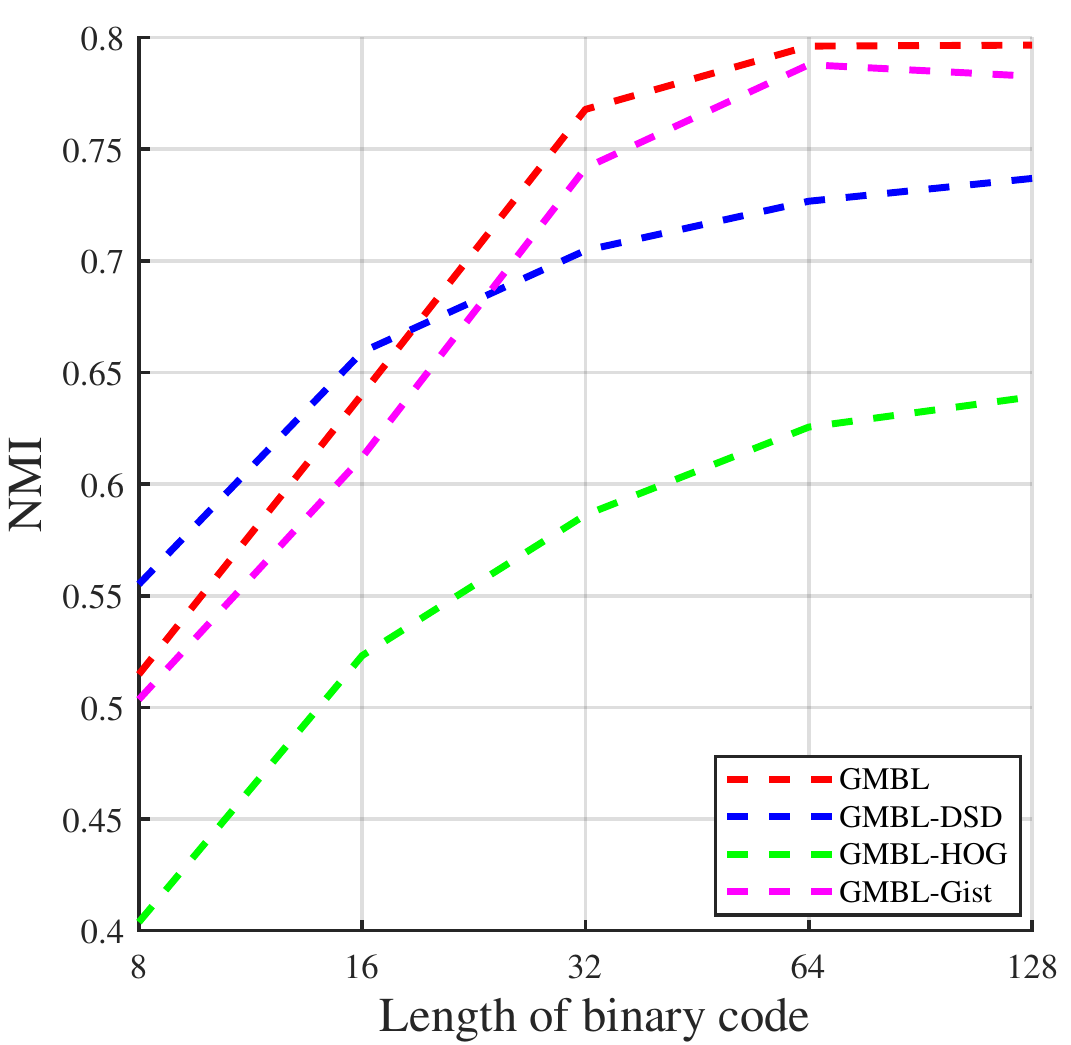}}
	\subfigure[]{
		\label{fig:subfig:b} 
		\includegraphics[width=3.8cm]{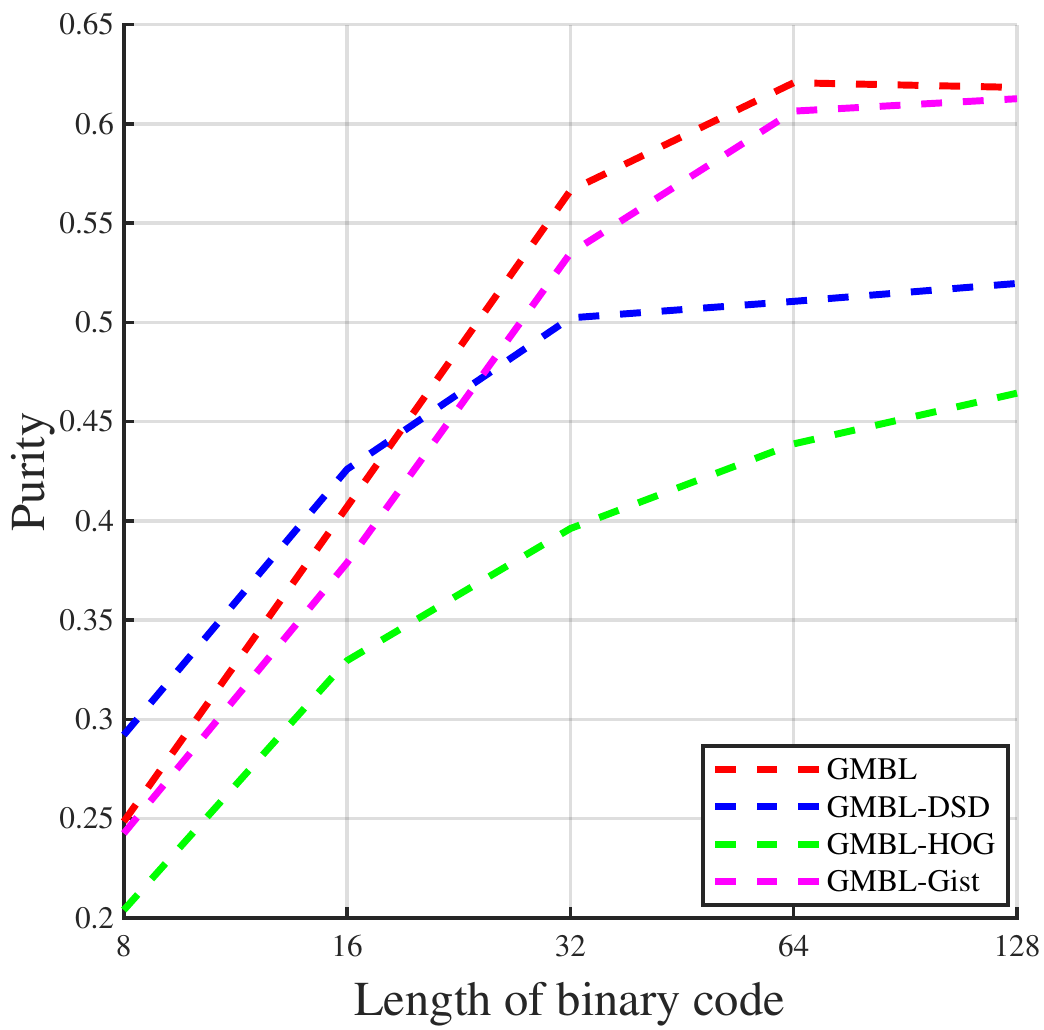}}
	\caption{multi-view vs single-view}	
	\label{Fig.8} 
\end{figure}

In order to verify more clearly, Fig.\ref{Fig.8} demonstrate the clustering performance of the algorithm in GMBL method of single views, evaluated by ACC, NMI and Purity respectively. We can notice that our multi-views method get higher results than the GMBL methods of single views. Fig.\ref{Fig.8} illustrates the clustering results of GMBL algorithm from different views on the Caltech101 and Coil-100 dataset. Multi-view GMBL still obtains higher results than a single view of GMBL when the clustering result of a certain view is remarkably efficient. In particular, with 128-bits, our multi-view method exceeds the best of single-view GMBL methods by more than 24\% and 67\% in terms of ACC, NMI and Purity, respectively. Thus, multi-view methods explore common cluster structure work better compared with single-view methods. 

\subsection{Comparison with state-of-art multi-view methods}

In this section, we present the detailed clustering results of three datasets in tables \ref{Table 5}, \ref{Table 6} and \ref{table 7}. In each table, the bold values illustrate the best clustering performance. These tables indicate that GMBL achieves excellent performance in four evaluation indexes of three datasets and is superior to other methods by Caltech101, clatech256 and NUS-WIDE-obj datasets. GMBL by learning discrete coding and real-value representation of multi-view clustering and has obtained encouraging results. In addition, even though the k-means clustering method concatenates all multiple views into a vector, it can not achieve efficient clustering performance. Because k-means clustering is essentially a single-view clustering method. In the experiment, the length of the hash code is set to 128-bits when compared with the real-valued multi-view clustering method.

\begin{table}[]
	\tiny	
	\centering	
	\caption{The clustering results on Caltech101 dataset}   \label{Table 5}
	\renewcommand\tabcolsep{5.0pt} 
	\begin{tabular}{@{}ccccccccccc@{}}
		\toprule
	Methods & k-means & SC     & Co-re-c & Co-re-s & AMGL   & Mul-NMF & MLAN   & BMVC   & HSIC   & GMBL   \\
		\midrule
		ACC     & 0.1331  & 0.1365 & 0.2670  & 0.2425  & 0.1350 & 0.2018  & 0.1807 & 0.2930 & 0.2578 &    \textbf{ 0.3070} \\
		NMI     & 0.3056  & 0.3269 & 0.4691  & 0.4683  & 0.2645 & 0.4089  & 0.2686 & 0.4900 & 0.3511 & \textbf{0.4982} \\
		Purity  & 0.2909  & 0.3187 & 0.4600  & 0.4694  & 0.1569 & 0.2300  & 0.3286 & 0.4907 & 0.3492 & \textbf{0.5008} \\
		F-score & 0.1895  & 0.0955 & 0.2295  & 0.1867  & 0.0319 & 0.1705  & 0.0481 & 0.2466 & 0.2502 & \textbf{0.2586} \\ 
	\bottomrule
\end{tabular}
\end{table}
Table \ref{Table 5} shows that the clustering results on the Caltech101 dataset. GMBL outperforms all the other methods on ACC, NMI, Purity, F-score and improves the baseline k-means more than 50\%. k-means is the worst method because the views are directly concatenated together and more noise will be introduced. The Co-regularization method expresses different views through co-regularization spectrum clustering to pursue a better clustering index, which is suitable for experiments with fewer perspectives and takes a longer time. Compared with real-value multi-view methods, GMBL has been improved significantly. The main reason is that BMVC learns the binary code of the different views in Hamming space and improve calculation efficiency.  However, the calculation of distance in Euclidean space by real-value multi-view method has low efficiency and high time cost. However, pursuing a similar matrix by graph-based clustering is time-consuming, Compared with the hash multi-view method, our calculation time is shorter.

\begin{table}[]
	\tiny	
	\centering	
	\caption{The clustering results on Caltech256 dataset}   \label{Table 6}
	\renewcommand\tabcolsep{5.0pt} 
	\begin{tabular}{@{}ccccccccccc@{}}
	\toprule
	Methods & k-means  & SC     & Co-re-c & Co-re-s & AMGL   & Mul-NMF & MLAN   & BMVC   & HSIC   & GMBL   \\ 
	\midrule
	ACC     & 0.1001  & 0.0924 & 0.1030  & 0.0738  & 0.0467 & 0.0713  & 0.0693 & 0.1028 & 0.0971 & \textbf{0.1049} \\
	NMI     & 0.1184  & 0.2764 & 0.2856  & 0.2467  & 0.1070 & 0.2272  & 0.0794 & 0.2915 & 0.2503 & \textbf{0.2949} \\
	Purity  & 0.1018  & 0.1339 & 0.1602  & 0.1070  & 0.0415 & 0.1119  & 0.0922 & 0.1428 & 0.1184 & \textbf{0.1475} \\
	F-score & 0.0804  & 0.0628 & 0.0727  & 0.0415  & 0.0466 & 0.0458  & 0.0224 & 0.0781 & 0.0719 &\textbf{ 0.0878} \\ 
	\bottomrule
\end{tabular}
\end{table}

For Caltech256 dataset, we randomly selected 20222 samples as experimental data, which extracts three features and 175 categories of pictures. The clustering results with different methods can be found in table \ref{Table 6}. GMBL method outperforms all the other methods on four evaluation metrics. Compared with the real-valued multi-view method, the hash method has obvious advantages in large datasets and takes the least time in clustering. Compared with other multi-view hash methods, the clustering performance of GMBL is improved. Therefore, it is very important to maintain the original spatial structure in the process of learning binary codes.

\begin{table}[]	
	\tiny	
	\centering	
	\caption{The clustering results on NUS-WIDE-obj dataset}   \label{table 7}	
	\renewcommand\tabcolsep{5.0pt} 
	\begin{tabular}{@{}ccccccccccc@{}}
		\toprule
		Methods & k-means & SC     & Co-re-c & Co-re-s & AMGL   & Mul-NMF & MLAN   & BMVC   & HSIC   & GMBL   \\ \midrule
		ACC     & 0.1459  & 0.1360 & 0.1521  & 0.1625  & 0.1281 & 0.1183  & 0.1554 & 0.1508 & 0.1621 & \textbf{0.1682} \\
		NMI     & 0.1415  & 0.1289 & 0.1505  & 0.1604  & 0.1362 & 0.1029  & 0.1199 & 0.1527 & 0.1625 & \textbf{0.1649} \\
		Purity  & 0.2576  & 0.2460 & 0.2816  & 0.2826  & 0.1484 & 0.1975  & 0.2604 & 0.2855 & 0.2790 & \textbf{0.2968} \\
		F-score & 0.1105  & 0.0840 & 0.1038  & 0.1018  & 0.1125 & 0.1128  & 0.1136  & 0.1090 & \textbf{0.1190} & 0.1126 \\ \bottomrule
	\end{tabular}
\end{table} 

We used the test set of the NUS-WIDE-obj dataset to complete the clustering task in table \ref{table 7} methods. Since some images in the dataset have multiple labels, the most representative label was adopted as ground-truth in our comparative experiment. It can be found that our method is superior to other methods in three indicators. The adoption of the similarity matrix is more conducive to the hidden structure of mining data, but the use of the hash method only improves significantly in time, and the evaluation result does not improve significantly. In addition, GMBL takes a lower evaluation index F-score than HSIC.

\subsection{ Visualization}

\begin{figure}[htb]	
	\centering	
	\subfigure[Orignal data]{	
		\label{fig:a} 
		\includegraphics[width=4cm]{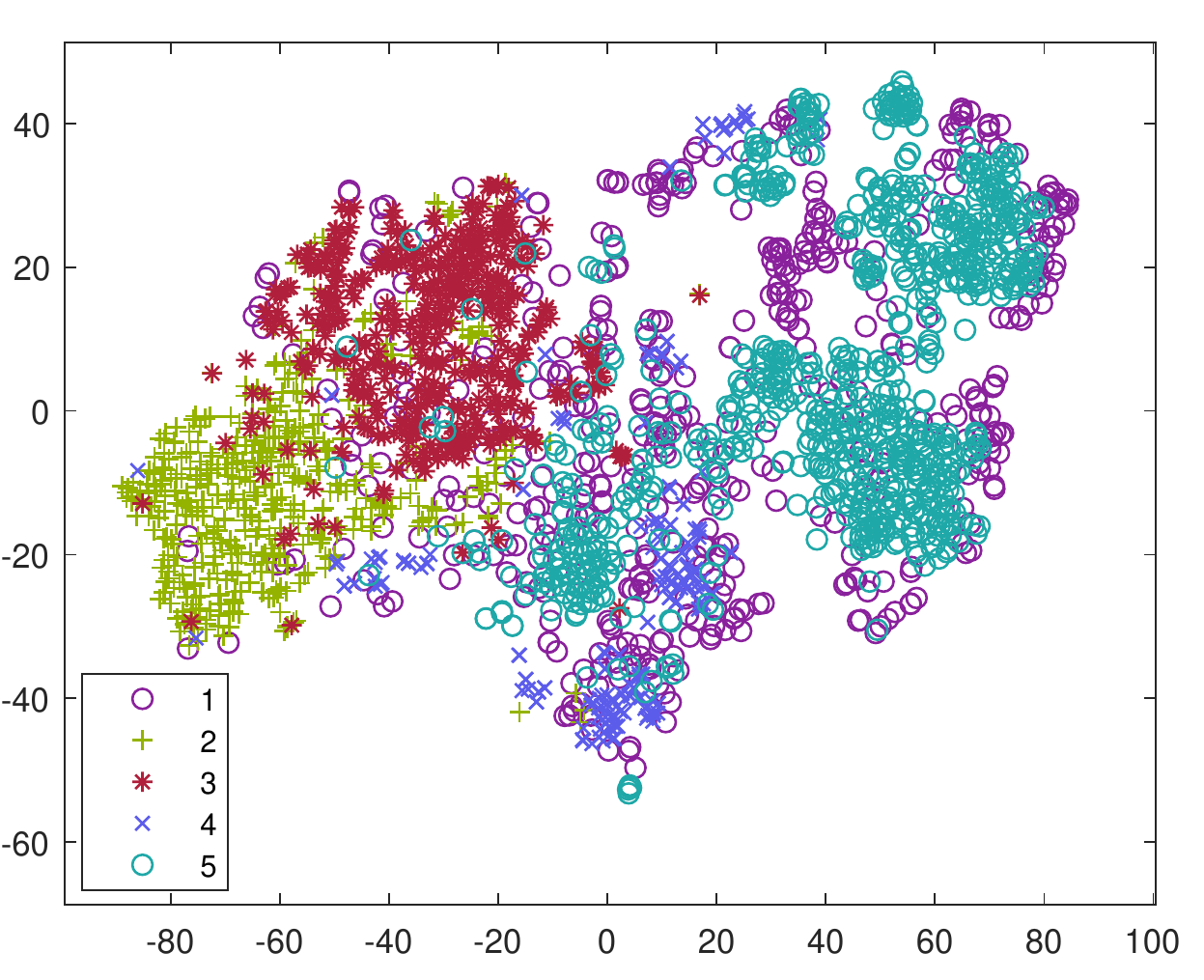}}
	\subfigure[Binary Codes of multiview]{
		\label{fig:subfig:b} 
		\includegraphics[width=4cm]{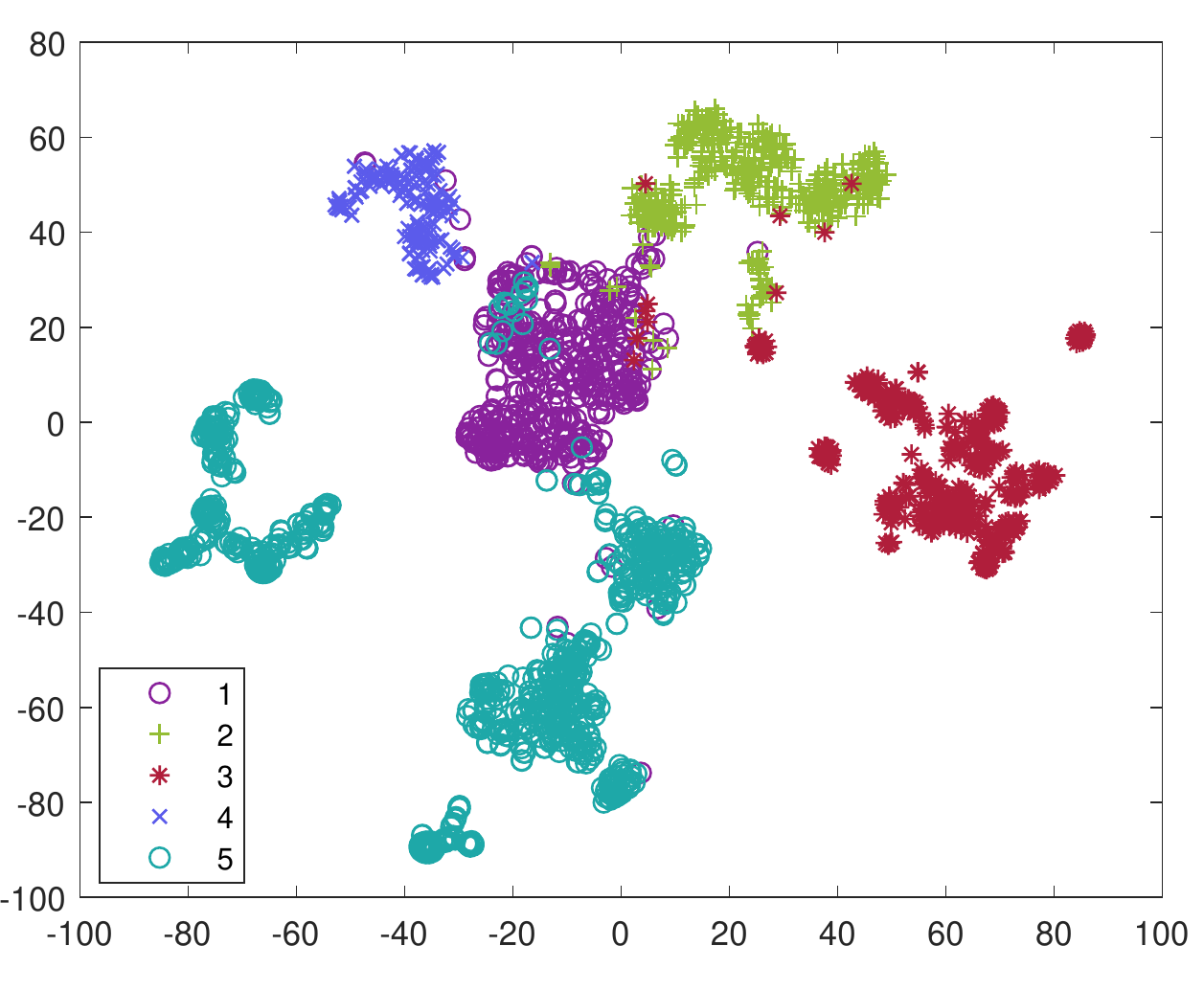}}
	\subfigure[Orignal data]{
		\label{fig:subfig:c} 
		\includegraphics[width=4cm]{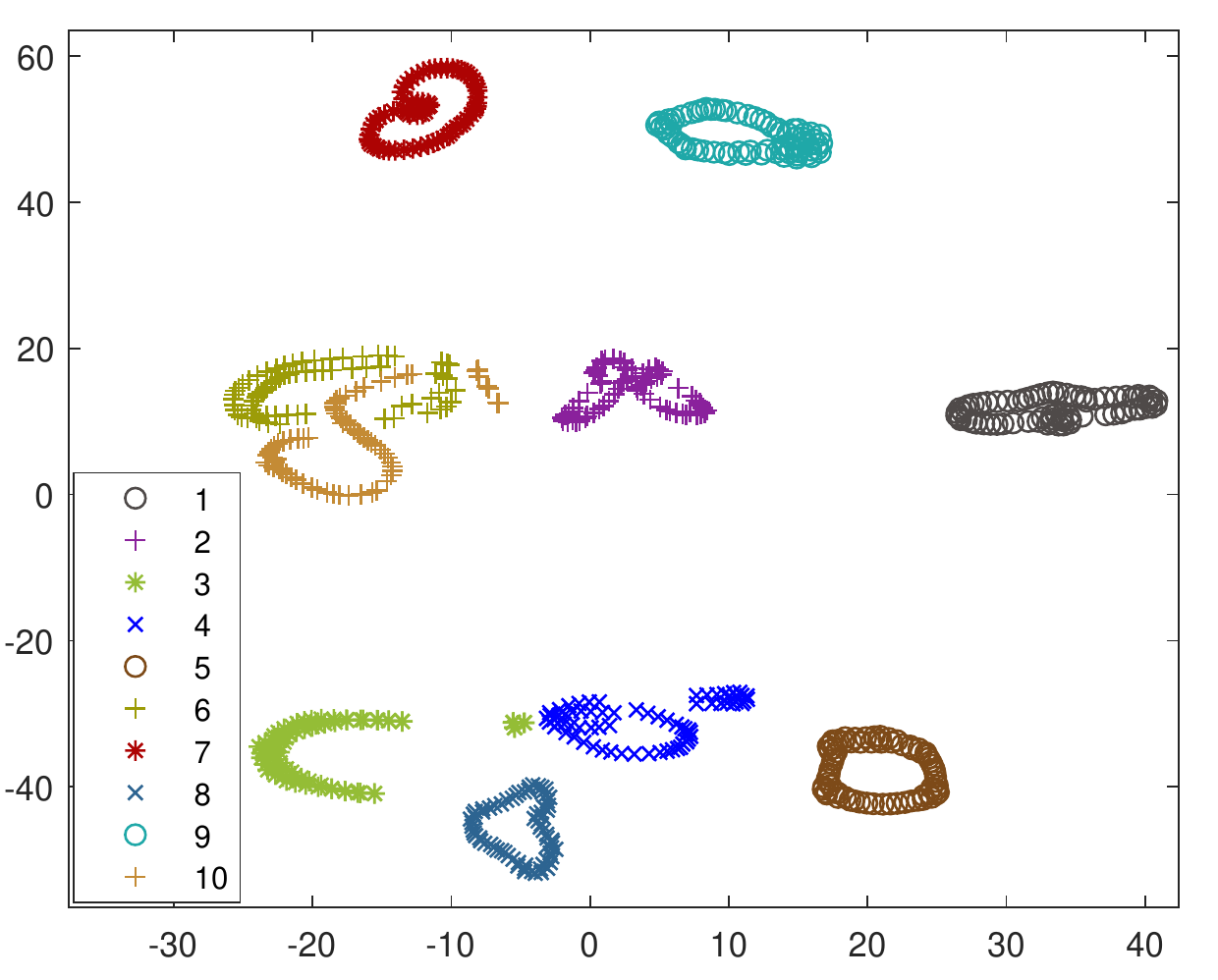}}
	\subfigure[Binary Codes of multiview]{
		\label{fig:subfig:c} 
		\includegraphics[width=4cm]{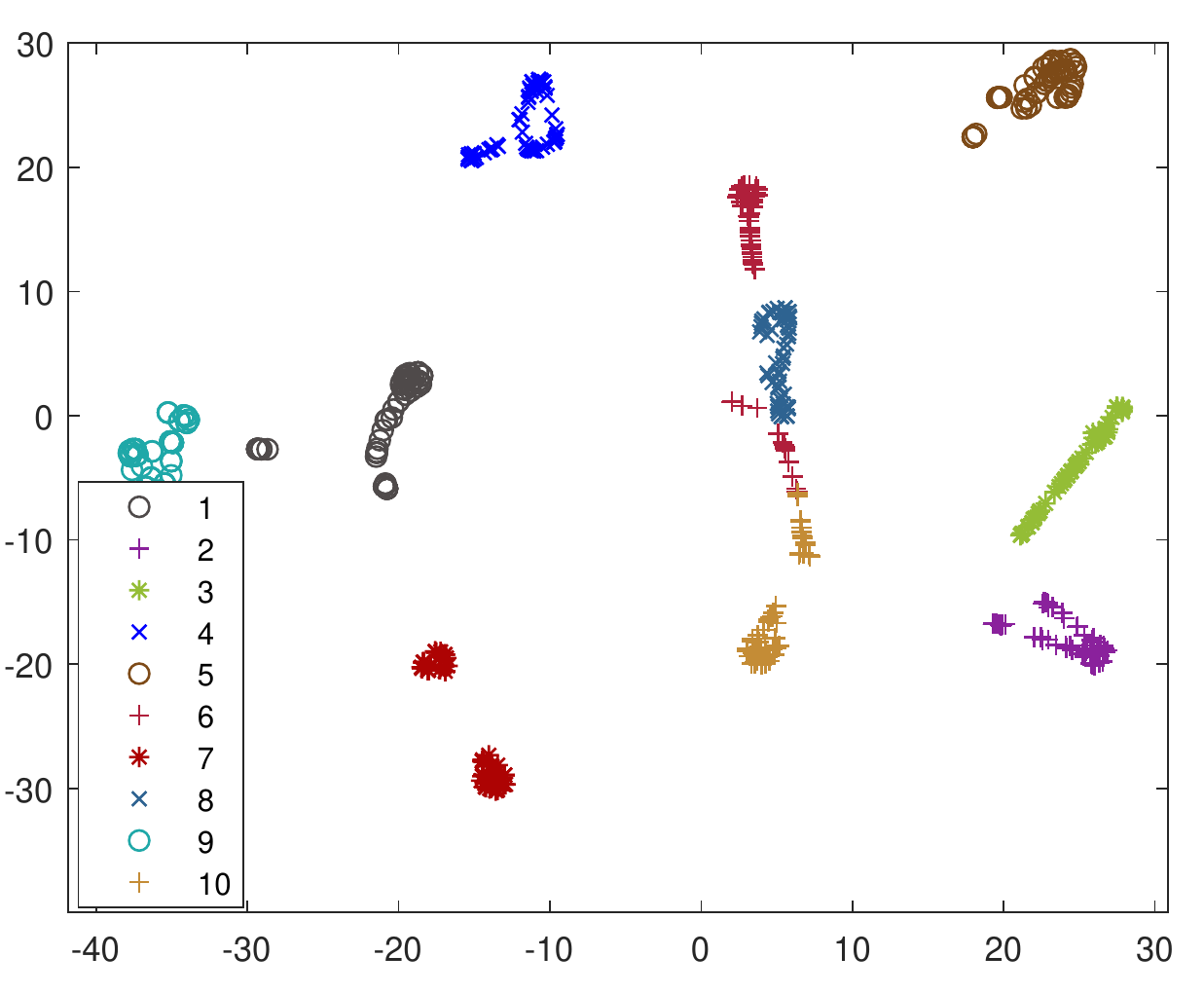}}
	\caption{Visualization of original features and binary encoding of clustering based on Caltech101 and Coil-100 datasets}	
	\label{Fig.9} 
\end{figure}

In Fig.\ref{Fig.9}, $a$ and $b$ are the visualization effects of binary codes and original data using t-SNE \cite{maaten2008visualizing} on the caltech101 dataset (we randomly selected 5 classes). The original data links all six features into a vector as input. In addition, $c$ and $d$ are the visualization effects of binary code and full connection raw data using t-SNE in the Coil-100 dataset (we randomly selected 10 classes). In Fig.\ref{Fig.9}, different colors geometric figures belong to various categories and the clustering results are well when the same kinds are adjacent to each other. We observed that the visualization of binary codes was more discriminating than the original data because each category in the graph was more scattered in the visualizations. 

All the experiments above can verify the excellent performance of our proposed GMBL. It can extend the Euclidean space measure method to binary code in Hamming space. Through the experiment results, GMBL is better than the real-value multi-view method exceeds in most situations. Compared with the hash method, the local structure of the data preserved by constructing the similarity matrix can effectively improve the clustering performance, which is stronger than the most hash algorithm.

\subsection {Convergence analysis}

\begin{figure}[htb]	
	\centering	
	\subfigure[Caltech101]{	
		\label{fig:a} 
		\includegraphics[width=5cm]{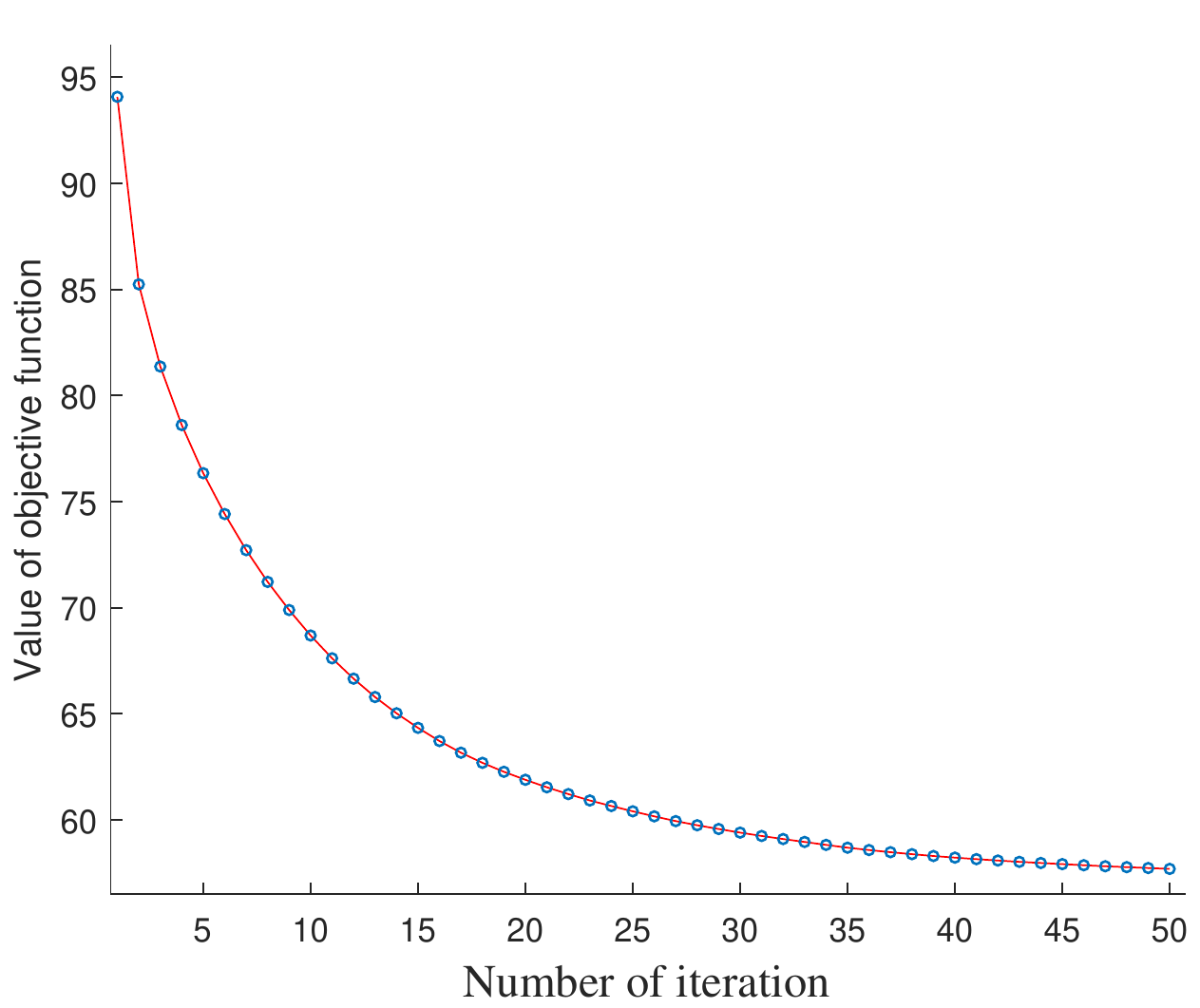}}
	\caption{Convergence Curve on Caltech101}	
	\label{Fig.10} 
\end{figure}

We adopted the Alternating iterative optimization method to iteratively update all the pending parameter matrix in our optimization problem. Fig.\ref{Fig.10} indicates the objective function values on the Caltech101 dataset. We observe that the values of our objective function on the dataset decrease rapidly in each iteration and access to a point. It can be identified that the constructed function is monotone and convergent and has minimum values.

\section{Conclusion}

In this paper, we propose a discrete hash coding algorithm based on graph clustering, named GMBL. GMBL learns efficient binary code, which can fully explore the original information of multi-view data and reduce the lack of information. With the Laplace similarity matrix, the proposed algorithm can be preserved in the local linear relationship of the original data and the multi-view binary clustering task can be well optimized. In addition, since various views contribute differently to cluster tasks, we assign weights to different views of adaptively giving to their contributions. In order to optimize the binary code, we adopt the alternating iteration method to directly optimize it instead of the loan constraint. It can be found that different from the traditional real-valued multi-view clustering method, the hashing clustering method can effectively reduce the experimental time. We evaluated our proposed framework on five multi-view datasets for experimental examination. The experiment demonstrated the superiority of our proposed method.

\section{Acknowledgements}
This work was supported in part by the National Natural Science Foundation of China Grant 61370142 and Grant 61272368, by the Fundamental Research Funds for the Central Universities Grant 3132016352, by the Fundamental Research of Ministry of Transport of P. R. China Grant 2015329225300, by Chinese Postdoctoral Science Foundation 3620080307, by the Dalian Science and Technology Innovation Fund 2018J12GX037 and Dalian Leading talent Grant, by
the Foundation of Liaoning Key Research and Development Program.

\bibliographystyle{elsarticle-num}
\bibliography{mybibfile}

\begin{thebibliography}{10}
\expandafter\ifx\csname url\endcsname\relax
  \def\url#1{\texttt{#1}}\fi
\expandafter\ifx\csname urlprefix\endcsname\relax\def\urlprefix{URL }\fi
\expandafter\ifx\csname href\endcsname\relax
  \def\href#1#2{#2} \def\path#1{#1}\fi

\bibitem{wang2015learning}
J.~Wang, W.~Liu, S.~Kumar, S.-F. Chang, Learning to hash for indexing big
  data—a survey, Proceedings of the IEEE 104~(1) (2015) 34--57.

\bibitem{bernabe2019efficient}
J.~A. Bernab{\'e}-D{\'\i}az, M.~del Carmen Legaz-Garc{\'\i}a, J.~M.
  Garc{\'\i}a, J.~T. Fern{\'a}ndez-Breis, Efficient, semantics-rich
  transformation and integration of large datasets, Expert Systems with
  Applications 133 (2019) 198--214.

\bibitem{dean2013fast}
T.~Dean, M.~A. Ruzon, M.~Segal, J.~Shlens, S.~Vijayanarasimhan, J.~Yagnik,
  Fast, accurate detection of 100,000 object classes on a single machine, in:
  Proceedings of the IEEE Conference on Computer Vision and Pattern
  Recognition, 2013, pp. 1814--1821.

\bibitem{fuentes2019topic}
G.~Fuentes-Pineda, I.~V. Meza-Ruiz, Topic discovery in massive text corpora
  based on min-hashing, Expert Systems with Applications.

\bibitem{sudharshan2019multiple}
P.~Sudharshan, C.~Petitjean, F.~Spanhol, L.~E. Oliveira, L.~Heutte, P.~Honeine,
  Multiple instance learning for histopathological breast cancer image
  classification, Expert Systems with Applications 117 (2019) 103--111.

\bibitem{ahmed2019hash}
T.~Ahmed, M.~Sarma, Hash-based space partitioning approach to iris biometric
  data indexing, Expert Systems with Applications 134 (2019) 1--13.

\bibitem{yang2017discrete}
R.~Yang, Y.~Shi, X.-S. Xu, Discrete multi-view hashing for effective image
  retrieval, in: Proceedings of the 2017 ACM on International Conference on
  Multimedia Retrieval, ACM, 2017, pp. 175--183.

\bibitem{datar2004locality}
M.~Datar, N.~Immorlica, P.~Indyk, V.~S. Mirrokni, Locality-sensitive hashing
  scheme based on p-stable distributions, in: Proceedings of the twentieth
  annual symposium on Computational geometry, ACM, 2004, pp. 253--262.

\bibitem{norouzi2011minimal}
M.~Norouzi, D.~M. Blei, Minimal loss hashing for compact binary codes, in:
  Proceedings of the 28th international conference on machine learning
  (ICML-11), Citeseer, 2011, pp. 353--360.

\bibitem{shen2015supervised}
F.~Shen, C.~Shen, W.~Liu, H.~Tao~Shen, Supervised discrete hashing, in:
  Proceedings of the IEEE conference on computer vision and pattern
  recognition, 2015, pp. 37--45.

\bibitem{gui2016supervised}
J.~Gui, T.~Liu, Z.~Sun, D.~Tao, T.~Tan, Supervised discrete hashing with
  relaxation, IEEE transactioSupervised ns on neural networks and learning
  systems 29~(3) (2016) 608--617.

\bibitem{gui2017fast}
J.~Gui, T.~Liu, Z.~Sun, D.~Tao, T.~Tan, Fast supervised discrete hashing, IEEE
  transactions on pattern analysis and machine intelligence 40~(2) (2017)
  490--496.

\bibitem{weiss2009spectral}
Y.~Weiss, A.~Torralba, R.~Fergus, Spectral hashing, in: Advances in neural
  information processing systems, 2009, pp. 1753--1760.

\bibitem{gong2012iterative}
Y.~Gong, S.~Lazebnik, A.~Gordo, F.~Perronnin, Iterative quantization: A
  procrustean approach to learning binary codes for large-scale image
  retrieval, IEEE Transactions on Pattern Analysis and Machine Intelligence
  35~(12) (2012) 2916--2929.

\bibitem{liu2014discrete}
W.~Liu, C.~Mu, S.~Kumar, S.-F. Chang, Discrete graph hashing, in: Advances in
  neural information processing systems, 2014, pp. 3419--3427.

\bibitem{shen2015hashing}
F.~Shen, C.~Shen, Q.~Shi, A.~Van~den Hengel, Z.~Tang, H.~T. Shen, Hashing on
  nonlinear manifolds, IEEE Transactions on Image Processing 24~(6) (2015)
  1839--1851.

\bibitem{wang2017effective}
Y.~Wang, X.~Lin, L.~Wu, W.~Zhang, Effective multi-query expansions:
  Collaborative deep networks for robust landmark retrieval, IEEE Transactions
  on Image Processing 26~(3) (2017) 1393--1404.

\bibitem{wu2018deep}
L.~Wu, Y.~Wang, X.~Li, J.~Gao, Deep attention-based spatially recursive
  networks for fine-grained visual recognition, IEEE transactions on
  cybernetics 49~(5) (2018) 1791--1802.

\bibitem{wang2015unsupervised}
Y.~Wang, W.~Zhang, L.~Wu, X.~Lin, X.~Zhao, Unsupervised metric fusion over
  multiview data by graph random walk-based cross-view diffusion, IEEE
  transactions on neural networks and learning systems 28~(1) (2015) 57--70.

\bibitem{wu2016exploiting}
L.~Wu, Y.~Wang, S.~Pan, Exploiting attribute correlations: A novel trace
  lasso-based weakly supervised dictionary learning method, IEEE transactions
  on cybernetics 47~(12) (2016) 4497--4508.

\bibitem{wu2018deep1}
L.~Wu, Y.~Wang, J.~Gao, X.~Li, Deep adaptive feature embedding with local
  sample distributions for person re-identification, Pattern Recognition 73
  (2018) 275--288.

\bibitem{wang2016iterative}
Y.~Wang, W.~Zhang, L.~Wu, X.~Lin, M.~Fang, S.~Pan, Iterative views agreement:
  An iterative low-rank based structured optimization method to multi-view
  spectral clustering, in: International Joint Conference on Artificial
  Intelligence (IJCAI), 2016.

\bibitem{dalal2005histograms}
N.~Dalal, B.~Triggs, Histograms of oriented gradients for human detection,
  2005.

\bibitem{ojala2002multiresolution}
T.~Ojala, M.~Pietik{\"a}inen, T.~M{\"a}enp{\"a}{\"a}, Multiresolution
  gray-scale and rotation invariant texture classification with local binary
  patterns, IEEE Transactions on Pattern Analysis \& Machine Intelligence~(7)
  (2002) 971--987.

\bibitem{rublee2011orb}
E.~Rublee, V.~Rabaud, K.~Konolige, G.~R. Bradski, Orb: An efficient alternative
  to sift or surf., in: ICCV, Vol.~11, Citeseer, 2011, p.~2.

\bibitem{xia2010multiview}
T.~Xia, D.~Tao, T.~Mei, Y.~Zhang, Multiview spectral embedding, IEEE
  Transactions on Systems, Man, and Cybernetics, Part B (Cybernetics) 40~(6)
  (2010) 1438--1446.

\bibitem{zhang2016flexible}
C.~Zhang, H.~Fu, Q.~Hu, P.~Zhu, X.~Cao, Flexible multi-view dimensionality
  co-reduction, IEEE Transactions on Image Processing 26~(2) (2016) 648--659.

\bibitem{wang2015robust}
Y.~Wang, X.~Lin, L.~Wu, W.~Zhang, Q.~Zhang, X.~Huang, Robust subspace
  clustering for multi-view data by exploiting correlation consensus, IEEE
  Transactions on Image Processing 24~(11) (2015) 3939--3949.

\bibitem{wu2018cycle}
L.~Wu, Y.~Wang, L.~Shao, Cycle-consistent deep generative hashing for
  cross-modal retrieval, IEEE Transactions on Image Processing 28~(4) (2018)
  1602--1612.

\bibitem{zhu2013linear}
X.~Zhu, Z.~Huang, H.~T. Shen, X.~Zhao, Linear cross-modal hashing for efficient
  multimedia search, in: Proceedings of the 21st ACM international conference
  on Multimedia, ACM, 2013, pp. 143--152.

\bibitem{ding2014collective}
G.~Ding, Y.~Guo, J.~Zhou, Collective matrix factorization hashing for
  multimodal data, in: Proceedings of the IEEE conference on computer vision
  and pattern recognition, 2014, pp. 2075--2082.

\bibitem{zhang2011composite}
D.~Zhang, F.~Wang, L.~Si, Composite hashing with multiple information sources,
  in: Proceedings of the 34th international ACM SIGIR conference on Research
  and development in Information Retrieval, ACM, 2011, pp. 225--234.

\bibitem{liu2015multiview}
L.~Liu, M.~Yu, L.~Shao, Multiview alignment hashing for efficient image search,
  IEEE Transactions on image processing 24~(3) (2015) 956--966.

\bibitem{shen2018multiview}
X.~Shen, F.~Shen, L.~Liu, Y.-H. Yuan, W.~Liu, Q.-S. Sun, Multiview discrete
  hashing for scalable multimedia search, ACM Transactions on Intelligent
  Systems and Technology (TIST) 9~(5) (2018) 53.

\bibitem{zhang2018binary}
Z.~Zhang, L.~Liu, F.~Shen, H.~T. Shen, L.~Shao, Binary multi-view clustering,
  IEEE transactions on pattern analysis and machine intelligence 41~(7) (2018)
  1774--1782.

\bibitem{shen2016fast}
F.~Shen, X.~Zhou, Y.~Yang, J.~Song, H.~T. Shen, D.~Tao, A fast optimization
  method for general binary code learning, IEEE Transactions on Image
  Processing 25~(12) (2016) 5610--5621.

\bibitem{gionis1999similarity}
A.~Gionis, P.~Indyk, R.~Motwani, et~al., Similarity search in high dimensions
  via hashing, in: Vldb, Vol.~99, 1999, pp. 518--529.

\bibitem{song2013effective}
J.~Song, Y.~Yang, Z.~Huang, H.~T. Shen, J.~Luo, Effective multiple feature
  hashing for large-scale near-duplicate video retrieval, IEEE Transactions on
  Multimedia 15~(8) (2013) 1997--2008.

\bibitem{jin2013density}
Z.~Jin, C.~Li, Y.~Lin, D.~Cai, Density sensitive hashing, IEEE transactions on
  cybernetics 44~(8) (2013) 1362--1371.

\bibitem{xia2015sparse}
Y.~Xia, K.~He, P.~Kohli, J.~Sun, Sparse projections for high-dimensional binary
  codes, in: Proceedings of the IEEE conference on computer vision and pattern
  recognition, 2015, pp. 3332--3339.

\bibitem{jiang2015scalable}
Q.-Y. Jiang, W.-J. Li, Scalable graph hashing with feature transformation, in:
  Twenty-Fourth International Joint Conference on Artificial Intelligence,
  2015.

\bibitem{zhang2018highly}
Z.~Zhang, L.~Liu, J.~Qin, F.~Zhu, F.~Shen, Y.~Xu, L.~Shao, H.~Tao~Shen,
  Highly-economized multi-view binary compression for scalable image
  clustering, in: Proceedings of the European Conference on Computer Vision
  (ECCV), 2018, pp. 717--732.

\bibitem{jetsadalak2018algorithm}
N.~Jetsadalak, O.~Suwunnamek, O.~T. Association, Y.~Sriwaranun, C.~Gan, M.~Lee,
  D.~Cohen, P.~Panpluem, M.~Putsakum, S.~Heaton, et~al., Algorithm as 136: A
  k-means clustering algorithm., Asian Journal of Scientific Research 12~(1)
  (2018) 480--510.

\bibitem{von2007tutorial}
U.~Von~Luxburg, A tutorial on spectral clustering, Statistics and computing
  17~(4) (2007) 395--416.

\bibitem{kumar2011co}
A.~Kumar, P.~Rai, H.~Daume, Co-regularized multi-view spectral clustering, in:
  Advances in neural information processing systems, 2011, pp. 1413--1421.

\bibitem{nie2016parameter}
F.~Nie, J.~Li, X.~Li, et~al., Parameter-free auto-weighted multiple graph
  learning: A framework for multiview clustering and semi-supervised
  classification., in: IJCAI, 2016, pp. 1881--1887.

\bibitem{liu2013multi}
J.~Liu, C.~Wang, J.~Gao, J.~Han, Multi-view clustering via joint nonnegative
  matrix factorization, in: Proceedings of the 2013 SIAM International
  Conference on Data Mining, SIAM, 2013, pp. 252--260.

\bibitem{nie2017multi}
F.~Nie, G.~Cai, X.~Li, Multi-view clustering and semi-supervised classification
  with adaptive neighbours, in: Thirty-First AAAI Conference on Artificial
  Intelligence, 2017.

\bibitem{gao2015multi}
H.~Gao, F.~Nie, X.~Li, H.~Huang, Multi-view subspace clustering, in:
  Proceedings of the IEEE international conference on computer vision, 2015,
  pp. 4238--4246.

\bibitem{cao2015diversity}
X.~Cao, C.~Zhang, H.~Fu, S.~Liu, H.~Zhang, Diversity-induced multi-view
  subspace clustering, in: Proceedings of the IEEE conference on computer
  vision and pattern recognition, 2015, pp. 586--594.

\bibitem{maaten2008visualizing}
L.~v.~d. Maaten, G.~Hinton, Visualizing data using t-sne, Journal of machine
  learning research 9~(Nov) (2008) 2579--2605.

\end{thebibliography}

\end{document}